\documentclass[10pt,twocolumn,letterpaper]{article}

\usepackage{cvpr}              

\usepackage{graphicx}
\usepackage{amsmath}
\usepackage{amssymb}
\usepackage{booktabs}
\usepackage{enumitem}
\setlist[itemize]{nosep}
\usepackage{bm}
\usepackage{makecell}

\usepackage[pagebackref,breaklinks,colorlinks]{hyperref}

\usepackage{marvosym}

\usepackage[capitalize]{cleveref}
\crefname{section}{Sec.}{Secs.}
\crefname{table}{Tab.}{Tabs.}

\def\cvprPaperID{1996} 
\def\confName{CVPR}
\def\confYear{2023}

\newcommand{\repName}{\emph{HairStep}\xspace}
\newcommand{\stranddataset}{\emph{HiSa}\xspace}
\newcommand{\depthdataset}{\emph{HiDa}\xspace}
\newcommand{\strandmetric}{\emph{HairSale}\xspace}
\newcommand{\depthmetric}{\emph{HairRida}\xspace}

\newcommand{\hairstep}{\mathbf{H}}
\newcommand{\strandmap}{\mathbf{O}}
\newcommand{\depthmap}{\mathbf{D}}
\newcommand{\pixel}{\mathbf{x}}

\newcommand{\orienttwodim}{\mathbf{O}_\mathrm{2D}}

\begin{document}

\title{\repName: Transfer Synthetic to Real Using Strand and Depth Maps\\ for Single-View 3D Hair Modeling }



\author{Yujian Zheng\textsuperscript{1,2} \quad Zirong Jin\textsuperscript{2} \quad  Moran Li\textsuperscript{3} \quad  Haibin Huang\textsuperscript{3} \\ \quad  Chongyang Ma\textsuperscript{3} \quad Shuguang Cui\textsuperscript{2,1} \quad Xiaoguang Han\textsuperscript{2,1*} \\
\textsuperscript{1}FNii, CUHKSZ\quad
\textsuperscript{2}SSE, CUHKSZ\quad
\textsuperscript{3}Kuaishou Technology\\
}

\newcommand{\cmt}[1]{\textcolor{red}{{[Comments: #1]}}}
\newcommand{\todo}[1]{\textcolor{blue}{{[todo: #1]}}}

\twocolumn[{%
\renewcommand\twocolumn[1][]{#1}%
\maketitle
\begin{center}
    \centering
    \captionsetup{type=figure}
    \includegraphics[width=\linewidth]{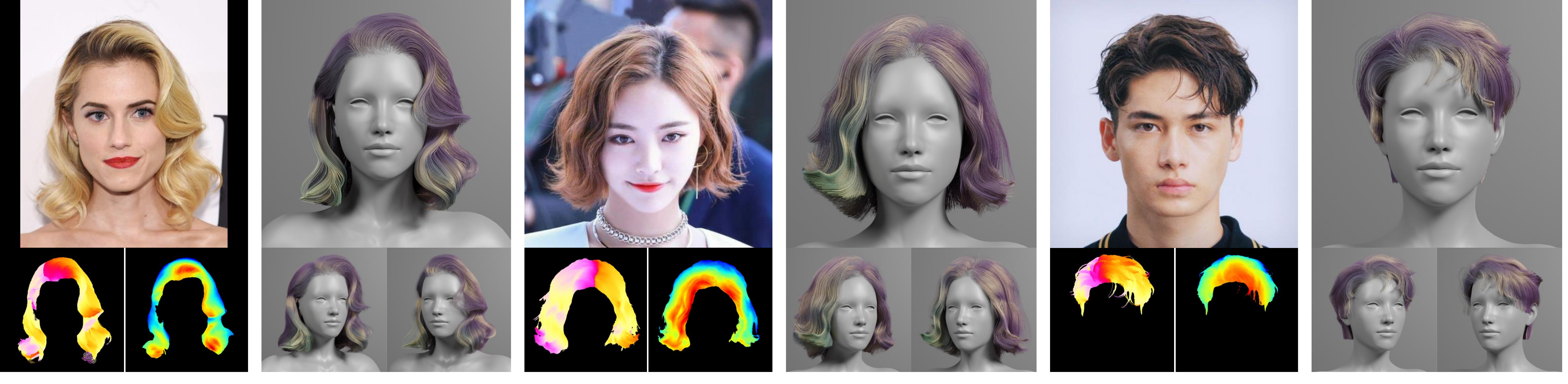}
    \captionof{figure}{Given a single portrait image, we first convert it to an intermediate representation \repName consisting of a strand map and a depth map (shown in the bottom left and right for each example), and then recover the corresponding 3D hair model at the strand level. Our \repName is capable to bridge the domain gap between synthetic and real data and achieves high-fidelity hair modeling results.} \label{fig:teaser}
\end{center}%
}]


\renewcommand{\thefootnote}{}
\footnotetext{\textsuperscript{*}Corresponding author: hanxiaoguang@cuhk.edu.cn}
\begin{abstract}
In this work, we tackle the challenging problem of learning-based single-view 3D hair modeling. Due to the great difficulty of collecting paired real image and 3D hair data, using synthetic data to provide prior knowledge for real domain becomes a leading solution. This unfortunately introduces the challenge of domain gap. Due to the inherent difficulty of realistic hair rendering, existing methods typically use orientation maps instead of hair images as input to bridge the gap. We firmly think an intermediate representation is essential, but we argue that orientation map using the dominant filtering-based methods is sensitive to uncertain noise and far from a competent representation. Thus, we first raise this issue up and propose a novel intermediate representation, termed as \textbf{HairStep}, which consists of a strand map and a depth map. It is found that HairStep not only provides sufficient information for accurate 3D hair modeling, but also is feasible to be inferred from real images. Specifically, we collect a dataset of 1,250 portrait images with two types of annotations. A learning framework is further designed to transfer real images to the strand map and depth map. It is noted that, an extra bonus of our new dataset is the first quantitative metric for 3D hair modeling. Our experiments show that HairStep narrows the domain gap between synthetic and real and achieves state-of-the-art performance on single-view 3D hair reconstruction.    

\end{abstract}

\section{Introduction}
\label{sec:intro}
High-fidelity 3D hair modeling is a critical part in the creation of digital human. 
A hairstyle of a person typically consists of about 100,000 strands~\cite{bao2018survey}. 
Due to the complexity, high-quality 3D hair model is expensive to obtain. 
Although high-end capture systems~\cite{luo2012multi,hu2014robust} are relatively mature, it is still difficult to reconstruct satisfactory 3D hair with complex geometries.

Chai \etal~\cite{chai2012single, chai2013dynamic} first present simple hair modeling methods from single-view images, which enable the acquisition of 3D hair more user-friendly. 
But these early systems require extra input such as user strokes. 
Moreover, they only work for visible parts of the hair and fail to recover invisible geometries faithfully.
Recently, retrieval-based approaches~\cite{hu2015single,chai2016autohair} reduce the dependency of user input and improve the quality of reconstructed 3D hair model. 
However, the accuracy and efficiency of these approaches are directly influenced by the size and diversity of the 3D hair database. 

Inspired by the advances of learning-based shape reconstruction, 3D strand models are generated by neural networks as explicit point sequences~\cite{zhou2018hairnet}, volumetric orientation field~\cite{saito20183d, zhang2019hair, shen2020deepsketchhair}, and implicit orientation field~\cite{wu2022neuralhdhair} from single-view input.
With the above evolution of 3D hair representations, the quality of recovered shape has been improved significantly. 
As populating pairs of 3D hair and real images is challenging~\cite{zhou2018hairnet}, existing learning-based methods~\cite{zhou2018hairnet,saito20183d,zhang2018modeling,shen2020deepsketchhair,wu2022neuralhdhair} are just trained on synthetic data before applying on real portraits. 
However, the domain gap between rendered images (from synthetic hair models) and real images has a great and negative impact on the quality of reconstructed results. 
3D hairstyles recovered by these approaches often mismatch the given images in some important details (e.g., orientation, curliness, and occlusion).

To narrow the domain gap between the synthetic data and real images, most existing methods~\cite{wu2022neuralhdhair,zhou2018hairnet,yang2019dynamic,zhang2019hair} take 2D orientation map~\cite{paris2004capture} as an intermediate representation between the input image and 3D hair model. 
However, this undirected 2D orientation map is ambiguous in growing direction and loses 3D hints given in the image. 
More importantly, it relies on image filters, which leads to noisy orientation maps. 
In this work, we re-consider the current issues in single-view 3D hair modeling and believe that it is necessary to find a more appropriate intermediate representation to bridge the domain gap between real and synthetic data. 
This representation should provide enough information for 3D hair reconstruction.
Also, it should be domain invariant and can be easily obtained from real image. 


To address the above issues, we propose \repName, a strand-aware and depth-enhanced hybrid representation for single-view 3D hair modeling. 
Motivated by how to generate clean orientation maps from real images, we annotate strand maps (i.e., directed 2D orientation maps) for real images via drawing well-aligned dense 2D vector curves along the hair.
With this help, we can predict directed and clean 2D orientation maps from input single-view images directly. 
We also need an extra component of the intermediate representation to provide 3D information for hair reconstruction. 
Inspired by depth-in-the-wild~\cite{chen2016single}, we annotate relative depth information for the hair region of real portraits. 
But depth learned from sparse and ordinal annotations has a non-negligible domain gap against the synthetic depth. 
To solve this, we propose a weakly-supervised domain adaptive solution based on the borrowed synthetic domain knowledge. 
Once we obtain the strand map and depth map, we combine them together to form \repName.
Then this hybrid representation will be fed into a network to learn 3D orientation field and 3D occupancy field of 3D hair models in implicit way. 
Finally, the 3D strand models can be synthesized from these two fields. The high-fidelity results are shown in~\cref{fig:teaser}. 
We name our dataset of hair images with strand annotation as \stranddataset and the one with depth annotation as \depthdataset for convenience.

Previous methods are mainly evaluated on real inputs through the comparison of the visual quality of reconstructed 3D hair and well-prepared user study.
This subjective measurement may lead to unfair evaluation and biased conclusion. 
NeuralHDHair~\cite{wu2022neuralhdhair} projects the growth direction of reconstructed 3D strands, and compares with the 2D orientation map filtered from real image. 
This is a noteworthy progress, but the extracted orientation map is noisy and inaccurate. 
Moreover, only 2D growing direction is evaluated and 3D information is ignored. 
Based on our annotations, we propose novel and objective metrics for the evaluation of single-view 3D hair modeling on realistic images. 
We render the recovered 3D hair model to obtain strand and depth map, then compare them with our ground-truth annotations.
Extensive experiments on our real dataset and the synthetic 3D hair dataset USC-HairSalon~\cite{hu2015single} demonstrate the superiority of our novel representation.

The main contributions of our work are as follows:
\begin{itemize}[leftmargin=*]
\item We first re-think the issue of the significant domain gap between synthetic and real data in single-view 3D hair modeling, and propose a novel representation \repName. Based on it, we provide a fully-automatic system for single-view hair strands reconstruction which achieves state-of-the-art performance.

\item We contribute two datasets, namely \stranddataset and \depthdataset, to annotate strand maps and depth for 1,250 hairstyles of real portrait images. 
This opens a door for future research about hair understanding, reconstruction and editing.

\item We carefully design a framework to generate \repName from real images. 
More importantly, we propose a weakly-supervised domain adaptive solution for hair depth estimation.

\item Based on our annotations, we introduce novel and fair metrics to evaluate the performance of single-view 3D hair modeling methods on real images. 
\end{itemize}

\section{Related Work}
\label{sec:related}

\begin{figure*}
\centering
\includegraphics[width=1.0\linewidth]{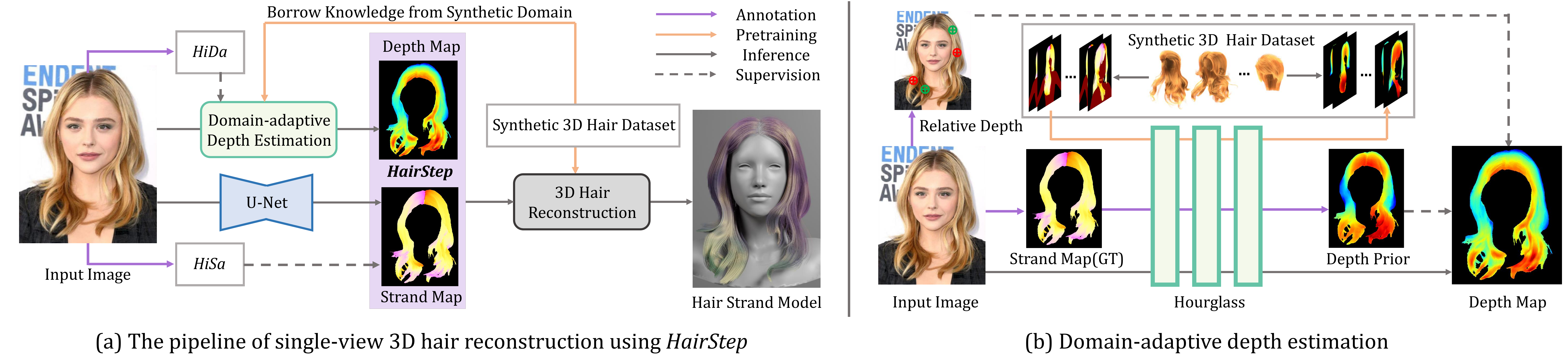}
\caption{Overview of our approach. 
(a) The pipeline of single-view 3D hair modeling with our novel representation \repName. 
We collect two datasets \stranddataset and \depthdataset, and propose effective approaches for \repName extraction from real images and finally realize high-fidelity 3D hair strand reconstruction. 
(b) Domain-adaptive depth estimation. 
We first pre-train the Hourglass on synthetic dataset, then generate depth priors as pseudo labels and finally obtain reasonable hair depth weakly-supervised by depth prior and annotated relative depth.
}
\label{fig:overview}
\end{figure*}


\paragraph{Single-view 3D hair modeling.}
It remains an open problem in computer vision and graphics to reconstruct 3D hair from a single-view input.
Compared with multi-view hair modeling~\cite{luo2012multi,zhang2018modeling,nam2019strand}, single-view methods~\cite{hu2015single,chai2012single,zhou2018hairnet,wu2022neuralhdhair} are more efficient and practical as multi-view approaches require carefully regulated environments and complex hardware setups.
The pioneering single-view based methods~\cite{hu2015single,chai2012single,chai2013dynamic,chai2016autohair} typically generate a coarse hair model based on a database first, and then use geometric optimization to approximate the target hairstyles. The effectiveness of these approaches relies on the quality of priors and the performance is less satisfactory for challenging input.

Recently, with the rapid development of deep learning, several methods~\cite{zhou2018hairnet,saito20183d,shen2020deepsketchhair,wu2022neuralhdhair} based on generative models have been proposed.
HairNet~\cite{zhou2018hairnet} takes the orientation map as the input to narrow the domain gap between real images and synthetic data, which enables the network to be trained with large-scale synthetic dataset.
Hair-VAE~\cite{saito20183d} adopts a variational autoencoder to generate hair models from single-view input.
Hair-GAN~\cite{zhang2019hair} introduces GAN based methods to the hair generation process.
However, the hair models reconstructed by these methods tend to be coarse and over-smoothed, mainly due to the limited capacity of 3D neural network.
To address this issue, NeuralHDHair~\cite{wu2022neuralhdhair} proposes a coarse-to-fine manner to obtain the high resolution 3D orientation fields and occupancy fields, enabling the GrowingNet to generate decent hair models.

\paragraph{Orientation maps for hair modeling.}
%
Due to the intrinsic elongated shapes of hair strands, it is intuitive to use 2D orientation maps and/or 3D orientation fields as intermediate representations to guide the modeling process.
Existing image-based hair modeling methods typically apply Gabor filters of different directions to the input portrait and compute the local 2D orientation to follow the direction with the maximum filtering response \cite{paris2004capture,paris2008hair}.
These 2D orientation maps are then converted into 3D orientation fields based on multi-view calibration information \cite{luo2012multi,luo2013wide,hu2014robust} or fed into neural network directly as auxiliary input for prediction of the 3D target hairstyle \cite{zhou2018hairnet,wu2022neuralhdhair,yang2019dynamic, zhang2019hair}.
However, 2D orientation maps based on image filtering operations suffer from input noise, which can be mitigated via additional smoothing or diffusion process at the expense of reduced accuracy~\cite{luo2012multi,luo2013wide}.
More importantly, these 2D orientation maps and 3D orientation fields do not distinguish between hair roots and tips from structure point of view. Addressing this kind of directional ambiguity requires additional input, such as user sketches~\cite{shen2020deepsketchhair} and physics based examples~\cite{hu2014robust}, which can be tedious or may not generalize well. 
Some methods\cite{tan2020michigan} for 2D hair image generation are also based on orientation map.


\paragraph{Depth map estimation.}
Many data-driven methods~\cite{karsch2014depth,hoiem2005automatic,saxena2008make3d,liu2015deep} using advanced techniques have achieved convincing performance on depth estimation.
However, these approaches rely on dense depth labeling~\cite{eigen2015predicting,silberman2012indoor,ladicky2014pulling,li2015depth}, which is inaccessible for hair strands.
Chen \etal~\cite{chen2016single} obviate the necessity of dense depth labeling by annotation of relative depth between sparse point pairs to help estimate depth map in the wild.
However, there is no existing work to estimate depth map specifically for hair strands.
Most 3D face or body reconstruction methods~\cite{saito2020pifuhd,tang2019neural,su2020robustfusion} only produce a coarse depth map of the hair region, which is far from enough for high-fidelity hair modeling.

\section{\repName Representation}
\label{sec:Repre}
The ideal way to recover 3D hair from single images via learning-based technique is to train a network which can map real images to the ground-truth 3D hair strands. 
But it is difficult and expensive to obtain ground-truth 3D hair geometries for real hair images~\cite{zhou2018hairnet}. 
\cite{saito20183d} can only utilize a retrieval-based method~\cite{hu2015single} to create pseudo 3D hair models. 
Networks trained on such data can not produce 3D hairstyles aligned with given images, because it is hard to guarantee the alignment of retrieved hair with the input image. 
Due to the inherent difficulty of realistic hair rendering, existing methods~\cite{wu2022neuralhdhair,zhou2018hairnet,yang2019dynamic,zhang2019hair} take orientation maps instead of hair images as input to narrow the domain gap between real and synthetic data. However, orientation map obtained by image filters suffers from uncertain noise and is far from a competent intermediate representation. 
Hence, a better one is needed to bridge the significant gap. 

We now formally introduce our novel representation \repName for single-view 3D hair modeling. The overview of our method is shown in \cref{fig:overview}.
We first give the definition of \repName in \cref{sec:definition}, then describe how to obtain it from real images in \cref{sec:extration_strand_map} and \cref{sec:domain_adaptive_depth_estimation}.
We describe how to use \repName for single-view 3D hair modeling in \cref{sec:recon}.

\subsection{Definition}
\label{sec:definition}
Given a target image, we define the corresponding representation \repName as $\hairstep=\{\strandmap, \depthmap\}$, where $\strandmap$ and $\depthmap$ are the strand map and the depth map, respectively. 
The strand map $\strandmap$ is formulated as an RGB image with a dimension of $W\times H\times 3$, where $W$ and $H$ are the width and the height of the target image.
The color at a certain pixel $\pixel$ on the strand map is defined as 
\begin{equation}
\label{eqn:01}
\begin{aligned}
\strandmap(\pixel) = (\textbf{M}(\pixel), \orienttwodim/2+0.5).
\end{aligned}
\end{equation}
We use the red channel to indicate the hair mask with a binary map $\textbf{M}$. 
We normalize the unit vector of projected 2D orientation $\orienttwodim$ of hair growth at pixel $\pixel$ and represent this growing direction in green and blue channels. 
The depth map $\depthmap$ can be easily defined as a $W\times H\times 1$ map where it represents the nearest distance of hair and the camera center in the camera coordinate at each pixel of hair region.
Visual examples of \repName are shown in~\cref{fig:teaser}. 

\paragraph{Difference with existing representations.} 
The existing 2D orientation map uses un-directed lines with two ambiguous directions~\cite{paris2004capture} to describe the pixel-level hair growing in the degree of 180 while our strand map can represent the direction in the degree of 360 (see~\cref{fig:strand_map_data} (d-e)). 
NeuralHDHair~\cite{wu2022neuralhdhair} attempts to introduce an extra luminance map to supplement the lost local details in the real image. 
Unfortunately, there is a non-negligible domain gap between the luminance of synthetic and real images. 
Because it is highly related to the rendering scenarios such as lighting and material. 
Compared to the luminance map, our hair depth map only contains geometric information, which helps to narrow the domain gap of the synthetic and real images. 


\subsection{Extraction of Strand Map}
\label{sec:extration_strand_map}
To enable learning-based single-view 3D hair modeling, \repName needs to be firstly extracted from both synthetic 3D hair data and real images for training and testing. 
For the synthetic data, we can easily obtain strand maps and depth maps from 3D strand models assisted by mature rendering techniques~\cite{liu2019soft}. 
But it is infeasible to extract strand maps from real images via existing approaches.
Thus, we use a learning-based approach and annotate a dataset \stranddataset to provide supervision.

\begin{figure}
	\centering
	\includegraphics[width=0.5\textwidth]  
	{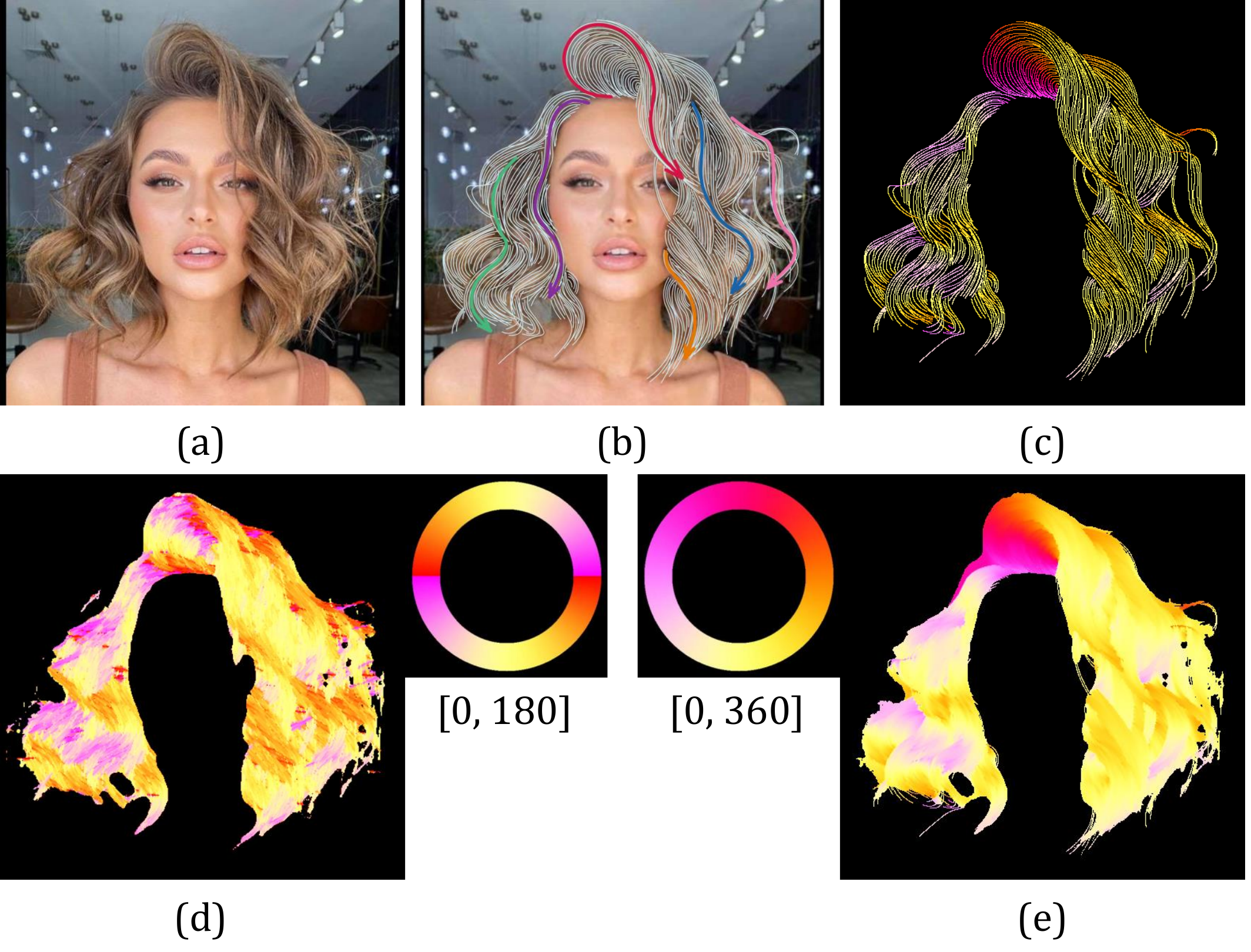}
	\caption{Obtaining strand map from vector strokes. (a) Portrait image. (b) Annotated vector strokes. (c) Colored strokes. (d) Orientation map extracted by Gabor filters. (e) Our strand map.}
	\label{fig:strand_map_data}
\end{figure}

\paragraph{\stranddataset dataset.}
We collect 1,250 clear portrait images with various hairstyles from the Internet. 
The statistics of the hairstyles, gender and race are given in the supplementary material.
We first hire artists to annotate dense 2D directional vector curves from the hair roots to the hair ends along the hair on the image (see the example in~\cref{fig:strand_map_data} (b)). 
On average, every hair image needs to cost about 1 hour of a skillful artist to draw about 300 vector curves. 
Once we obtain the dense strokes of the hair region, we convert them to a stroke map colored by the definition of~\cref{eqn:01}, as shown in~\cref{fig:strand_map_data} (c). 
At last, we interpolate the colorized strokes within the mask of hair to obtain the ground-truth strand map (\cref{fig:strand_map_data} (e)) of a given image.
Thanks to the dense annotation, the holes are simple to be filled with ignorable loss of details. 
Compared with the undirectional orientation map extracted by Gabor filters (\cref{fig:strand_map_data} (d)), our strand map is clean and can represent the growing direction without ambiguity. 


\paragraph{Strand map prediction.}
We consider the extraction of a strand map from a real image as an image-to-image translation task.
We find that simply using an U-Net~\cite{ronneberger2015u} can already achieve satisfactory results. 
Following standard settings, we use a pixel-wise $L_1$ loss and a perceptual loss against the ground-truth strand map $\textbf{O}$, which is formulated as
\begin{equation}
\label{eqn:02}
\begin{aligned}
L_{strand}= &\frac{1}{C \cdot \sum \textbf{M}}   \left \| \hat{\textbf{O}} - \textbf{O}\right \|_1   +\\   & \alpha \cdot   \frac{1}{W_j H_j C_j}   \left \| \phi_j(\hat{\textbf{O}})-\phi_j (\textbf{O})  \right \|_2^2,
\end{aligned}
\end{equation}
where $\hat{\textbf{O}}$ represents the predicted strand map and $C$ represents the channel number of orientation map. 
The function $\phi_j(\cdot)$ represents the former $j$ layers of pretrained VGG-19~\cite{simonyan2014very} and we set $j$ to be 35. $W_j$, $H_j$ and $C_j$ represent the shapes of output feature from $\phi_j(\cdot)$. 

\subsection{Domain-Adaptive Depth Estimation}
\label{sec:domain_adaptive_depth_estimation}

It is not trivial to obtain the depth of hair from real images, because we cannot directly acquire the ground-truth depth annotation. 
Inspired by depth-in-the-wild~\cite{chen2016single}, we annotate relative depth for the hair region of real images as weak labels. 
However, only constrained by the ordinal depth information, networks tend to generate unnatural depth maps. 
There is an obvious domain gap between learned depth from weak label and the synthetic depth used in the training, which leads to poor generalization when applying the trained model on real domain. 
Following the popular framework of domain adaptation based on pseudo labels~\cite{zheng2018t2net,zhao2019geometry,song2020learning,liang2021domain,zhang2021prototypical,fan2022self}, we propose a domain-adaptive depth estimation method to reduce the gap of depth maps from real and synthetic data (see~\cref{fig:overview}). 

\paragraph{\depthdataset dataset.}
We annotate depth relations for randomly selected pixel pairs in the hair region of each image among 1,250 portraits in \stranddataset. 
Different from depth-in-the-wild that only selects one pair per image, we annotate more than 140 pairs on average for each portrait which can give a more accurate and dense prediction.
We first generate super-pixels within the hair region according to the ratio of the area of hair and face. We then randomly sample pixel pairs from all adjacent super-pixels and finally generate 177,074 pixel pairs in total for 1,250 real images. 
Two points in a pair are colored to red and blue, respectively. 
A QA program is designed to annotate the ordinal depth by showing one pair on the image each time and ask ``which point in a pair of sampled pixels looks like closer to you, Red Point, Blue Point, or Hard to Tell?'', following~\cite{chen2016single}. 
12 well-trained workers are invited to annotate, which are split into three groups to ensure that every selected pair has been annotated three times by different groups. 
Finally 129,079 valid answers are collected (all groups give a certain relative depth, \ie red or blue, and agree with each other). 
Our samplings takes a median of 4.6 seconds for a worker to decide, and three groups agree on the relative depth 72.9\% of the time.

\paragraph{Learning depth map.}
We follow~\cite{chen2016single} to directly learn the mapping between the input image $\textbf{I}$ and the output dense depth map $\textbf{D}_r$ of the hair region through a Hourglass network~\cite{newell2016stacked}, which is weakly supervised by our annotations. 
To train the network using ordinal labels of depth, we need a loss function that encourages the predicted depth map to agree with the ground-truth relations.
We have found that the margin-ranking loss used in~\cite{yang2016deep,lin2018hallucinated,zhang2019ranksrgan} works well in our task:
\begin{equation}
\label{eqn:03}
L_{rank}= \frac{1}{N}\sum_{i=1}^{N} \max(0,-(\textbf{D}_r(p_1^i)-\textbf{D}_r(p_2^i))\cdot r^i + \varepsilon), 
\end{equation}
where $p^i_1$ and $p^i_2$ are pixels of the $i_{th}$ annotated pair $p^i$, $r^i$ is the ground-truth label which is set to 1 if $p^i_1$ is closer otherwise -1. 
$N$ represents the total number of sampled pairs in an image.
$\varepsilon$ is set to be 0.05, which gives a control to the difference of the depth values in $p_1^i$ and $p_2^i$.

\paragraph{Domain adaptation.}
Although the ordinal label can provide local depth variation, it is a weak constraint which introduces ambiguity and leads to uneven solutions. 
The predicted depth map is usually unnatural and full of jagged artifacts (see the side views in~\cref{fig:cpmpare_depth}). 
Applying this kind of depth to hair modeling often leads to coarse and noisy 3D shapes. 
To address above issues, we propose a weakly supervised domain-adaptive solution for hair depth estimation. 
We believe the knowledge borrowed from synthetic domain can help improve the quality of the learned depth. 

Network trained with ordinal labels can not sense the absolute location, size and range of depth. 
The predicted depth has a major domain gap comparing to the synthetic depth map used in the training of 3D hair modeling. 
To give a constraint of the synthetic domain, we first train a network $Depth_{syn}$ to predict depth maps from strand maps on synthetic dataset by minimizing the $L_1$ distance between the prediction and the synthetic ground-truth. 
Then we input ground-truth strand maps of real images to $Depth_{syn}$ to query pseudo labels $\bar{\textbf{D}}$ as depth priors. 
Note that directly applying this pseudo depth map to 3D hair modeling is not reasonable, because taking strand map as input can not provide adequate 3D information to the network. 
Jointly supervised by the depth prior and the weak-label of relative depth annotation, we predict decent depth maps which is not only natural-looking but preserves local relations of depth ranking. 
The loss function of the domain adaptive depth estimation is consisting of two parts, \ie, an $L_1$ loss against the pseudo label and the ranking loss defined in~\cref{eqn:03}: 
\begin{equation}
\label{eqn:04}
\begin{aligned}
L_{depth}= \beta \cdot \left \| \textbf{D}_r - \bar{\textbf{D}}\right \|_1 + L_{rank}.
\end{aligned}
\end{equation}

\section{Single-View 3D Hair Modeling}
\label{sec:recon}
Given the \repName representation of a single-view portrait image, we further recover it to a strand-level 3D hair model. 
In this section, we first illustrate the 3D implicit hair representation, then describe the procedure of the reconstruction for hair strands.

\subsection{3D Hair Representation}
Following NeuralHDHair~\cite{wu2022neuralhdhair}, which is considered to be state-of-the-art in single-view hair modeling, we use implicit occupancy field and orientation field to represent 3D hair model in the canonical space of a standard scalp. 
The value of a point within the occupancy field is assigned to 1 if it is inside of the hair volume and is set to 0 otherwise. 
The attribute of a point in orientation field is defined as the unit 3D direction of the hair growth.
The orientations of points outside of the hair volume are defined as zero vectors. 

We use the same approach as~\cite{saito20183d} to extract the hair surface. 
During training, we sample large amount of points to form a discrete occupancy field. 
The sampling strategy follows~\cite{saito2019pifu} which samples around the mesh surface randomly and within the bounding box uniformly with a ratio of 1:1. 
For the orientation field, we calculate unit 3D orientations for dense points along more than 10k strands each model. 


\subsection{Strand Generation}
To generate 3D stands from \repName, we first train a neural network NeuralHDHair* following the method described by Wu \etal~\cite{wu2022neuralhdhair}. 
Taking our \repName as input, the network can predict the implicit occupancy field and orientation field representing the target 3D hair model. 
Then we synthesis the hair strands adopting the growing method in~\cite{shen2020deepsketchhair} from the hair roots of the standard scalp.

The code of NeuralHDHair~\cite{wu2022neuralhdhair} has not been released yet and our own implementation NeuralHDHair* preserves the main pipeline and the full loss functions of NeuralHDHair, but has two main differences from the original NeuralHDHair.
First, we do not use the sub-module of luminance map. 
The luminance has the potential to provide more hints for hair reconstruction, but suffers from the apparent domain gap between synthetic and real images, since it is highly related to the lighting. 
We attempt to apply the luminance map to the NeuralHDHair*, but it can only bring minor improvement. 
Second, we discard the GrowingNet of NeuralHDHair, since our work focuses on the quality of the reconstruction results instead of efficiency, while the GrowingNet is designed to accelerate the conversion from 3D implicit fields to hair strands. 
It maintains the same growth performance comparing to the traditional hair growth algorithm of~\cite{shen2020deepsketchhair}, which is reported in~\cite{wu2022neuralhdhair}.

\section{Experiments}
\label{sec:exp}
\subsection{Datasets}
We train the proposed method on USC-HairSalon~\cite{hu2015single} which is a publicly accessible 3D hairstyle database consisting of 343 synthetic hair models in various styles. 
We follow~\cite{zhou2018hairnet} to augment 3D hair data and select 3 random views each hair to generate corresponding strand maps and depth maps to form our \repName.
As for our real datasets \stranddataset and \depthdataset, we use 1,054 images with the resolution of $512\times512$ for training and 196 for testing. During training, we augment images and annotations by random rotating, scaling, translating and horizontally flipping. 



\subsection{Evaluation Metrics}
Based on \stranddataset and \depthdataset, we propose two novel and fair metrics, i.e., \strandmetric and \depthmetric, to evaluate single-view 3D hair modeling results.
We render the reconstructed 3D hair model to obtain strand map $\textbf{O}_r$ and depth map $\textbf{D}_r$, then compare them with our ground-truth annotations $\textbf{O}_{gt}$ and $\textbf{D}_{gt}$. 
Also, these two metrics can be applied to the evaluation on \repName extraction. 

\paragraph{\strandmetric.}
We first compute the mean angle error of growing direction called \strandmetric on rendered strand map, which ranges from 0 to 180. We define the \strandmetric as 
\begin{equation}
\label{eqn:05}
\begin{aligned}
\strandmetric = \frac{1}{K}\sum_{\textbf{x}^i \in U} \arccos( \mathcal V(\textbf{O}_r(\textbf{x}^i)) \cdot \mathcal V(\textbf{O}_{gt}(\textbf{x}^i)) ),
\end{aligned}
\end{equation}
where $U$ is the intersected region of rendered mask and the ground-truth. $K$ is the total number of pixels in $U$. $\mathcal V(\textbf{O}_r(\textbf{x}^i))$ converts the color at pixel $\textbf{x}^i$ of strand map $\textbf{O}_r$ to an unit vector representing the growing direction. 

\paragraph{\depthmetric.}
The \strandmetric only test the degree of matching in 2D. 
We also need a metric \depthmetric to measure the relative depth accuracy on \depthdataset, which is defined as 
\begin{equation}
\label{eqn:06}
\begin{aligned}
\depthmetric = \frac{1}{Q}\sum_{i=1}^{Q} \max (0, r^i \cdot \mathrm{sign}(\textbf{D}_r(p^i_1)-\textbf{D}_r(p^i_2))).
\end{aligned}
\end{equation}
Note that we also calculate \depthmetric in the intersected region of rendered mask and the ground-truth. 
In addition, we provide the statistics of IoU for reference. 

As for the evaluation of synthetic data, we follow~\cite{wu2022neuralhdhair} to compute the precision for occupancy field while using the $L_2$ error for orientation field.

\subsection{Evaluation on HairStep Extraction}
\begin{figure}
	\centering
	\includegraphics[width=0.9\linewidth]  
	{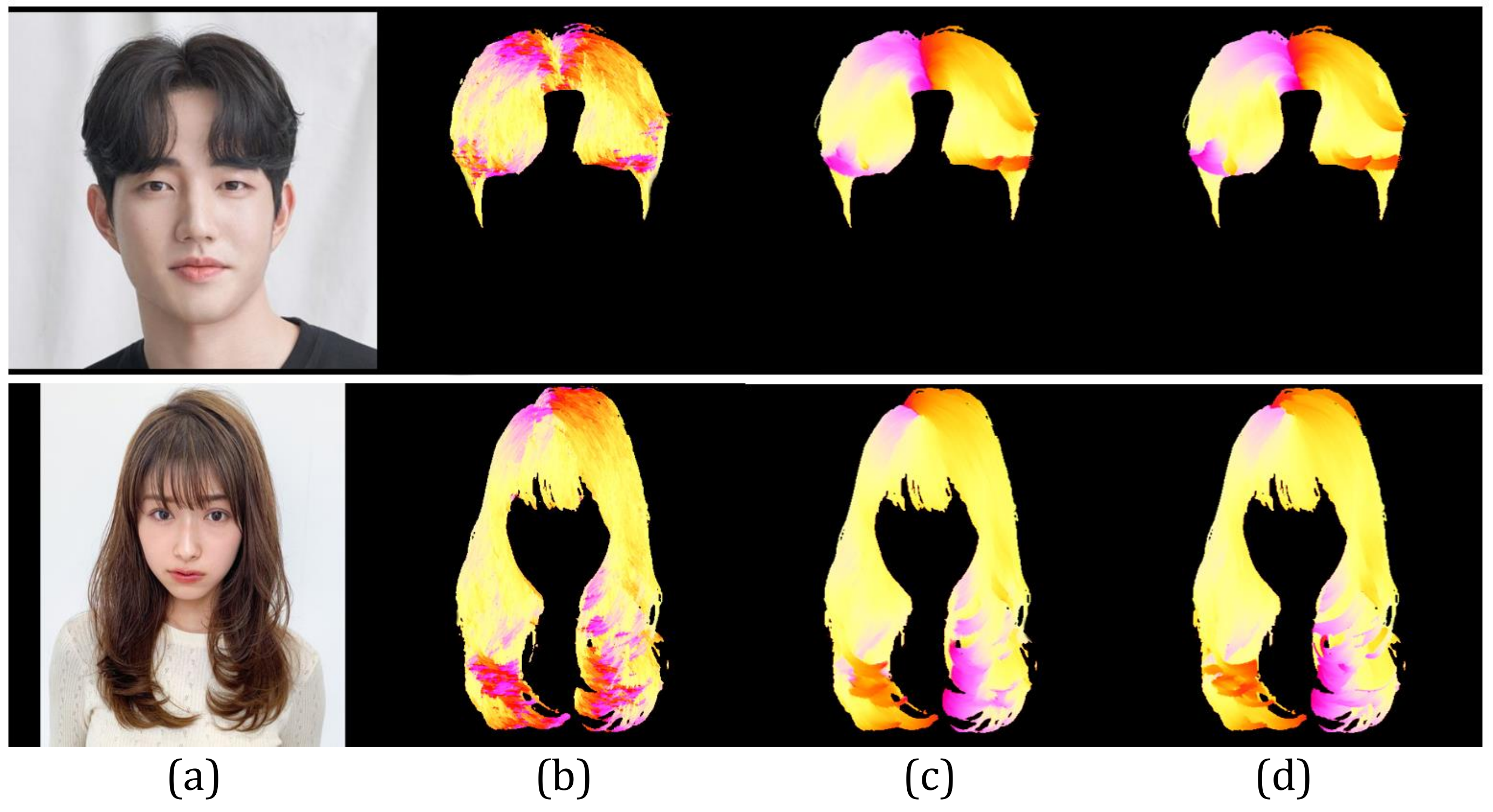}
	\caption{Qualitative comparisons on orientation/strand maps. (a) Input images; (b) undirected orientation maps from Gabor filters; (c) strand maps from our method; (d) ground-truth strand maps.}
	\label{fig:compare_orien}
\end{figure}

We first evaluate the effectiveness of our \repName extraction method from real images. 
We found that simply applying a U-Net can already generate clean strand maps while Gabor filters suffer from uncertain noises (see~\cref{fig:compare_orien}). 
The \strandmetric computed on our predicted strand map is 12.3. 
As Gabor filters can only produce undirected orientation map, we convert the strand map to undirected map to calculate \strandmetric quantitatively for fair comparison. 
The \strandmetric on our results and Gabor's are 14.2 and 18.4 in undirected way, where our method performs 22.8\% better.
It's worth mentioning, the errors in undirected way are larger than in directed way, since the ambiguous bi-directed orientation leads to a worse measurement.


\begin{figure}
	\centering
	\includegraphics[width=0.9\linewidth]  
	{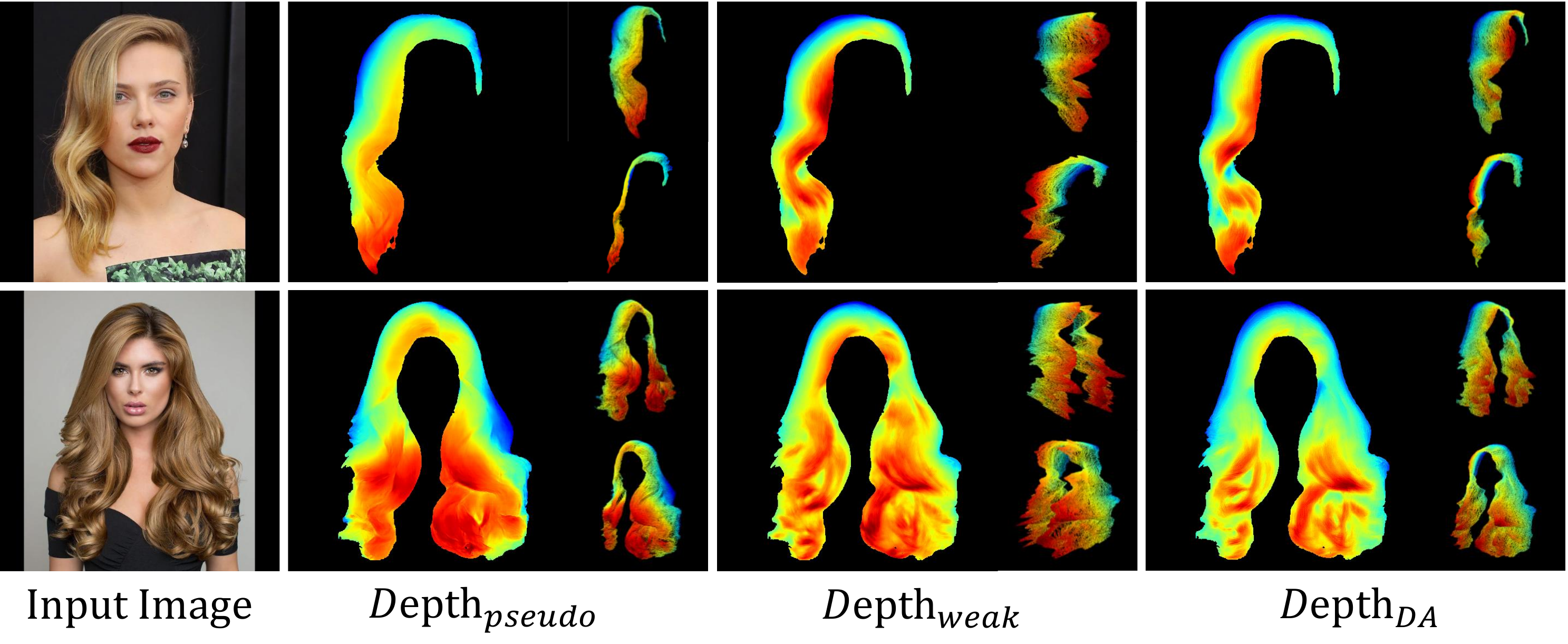}
	\caption{Qualitative comparisons on depth estimation.}
	\label{fig:cpmpare_depth}
\end{figure}

We evaluate the depth estimation using two metrics:  
the \depthmetric and a $L_1$ error against pseudo label (w/ or w/o normalization) to measure the difference between predicted depth and the synthetic prior. 
We compare the results of our domain-adaptive method $Depth_{DA}$ with the pseudo label $Depth_{pseudo}$ from synthetic domain, as well as the results of method only weakly supervised by ordinal label $Depth_{weak}$. 
The \depthmetric for $Depth_{pseudo}$, $Depth_{weak}$ and $Depth_{DA}$ are 80.47$\%$, 85.17$\%$ and 85.20$\%$, respectively. 
$L_1$ error against pseudo label (w/ ot w/o normalization) for $Depth_{weak}$ and $Depth_{DA}$ are 0.2470/3.125 and 0.1768/0.1188. 
Qualitative comparisons with different views of point cloud converted from depth maps are also shown in~\cref{fig:cpmpare_depth}.
The quantitative and qualitative give the same conclusion that our $Depth_{DA}$ is more competent to balance the local details of depth and the similarity of global shape to the synthetic prior. 
But the $Depth_{weak}$ is unnatural and full of serration-like artifacts.
$Depth_{pseudo}$ suffers from the flat geometry, because the strand map can not provide strong 3D hints. 


\subsection{Comparisons}
\begin{figure}
	\centering
	\includegraphics[width=\linewidth]  
	{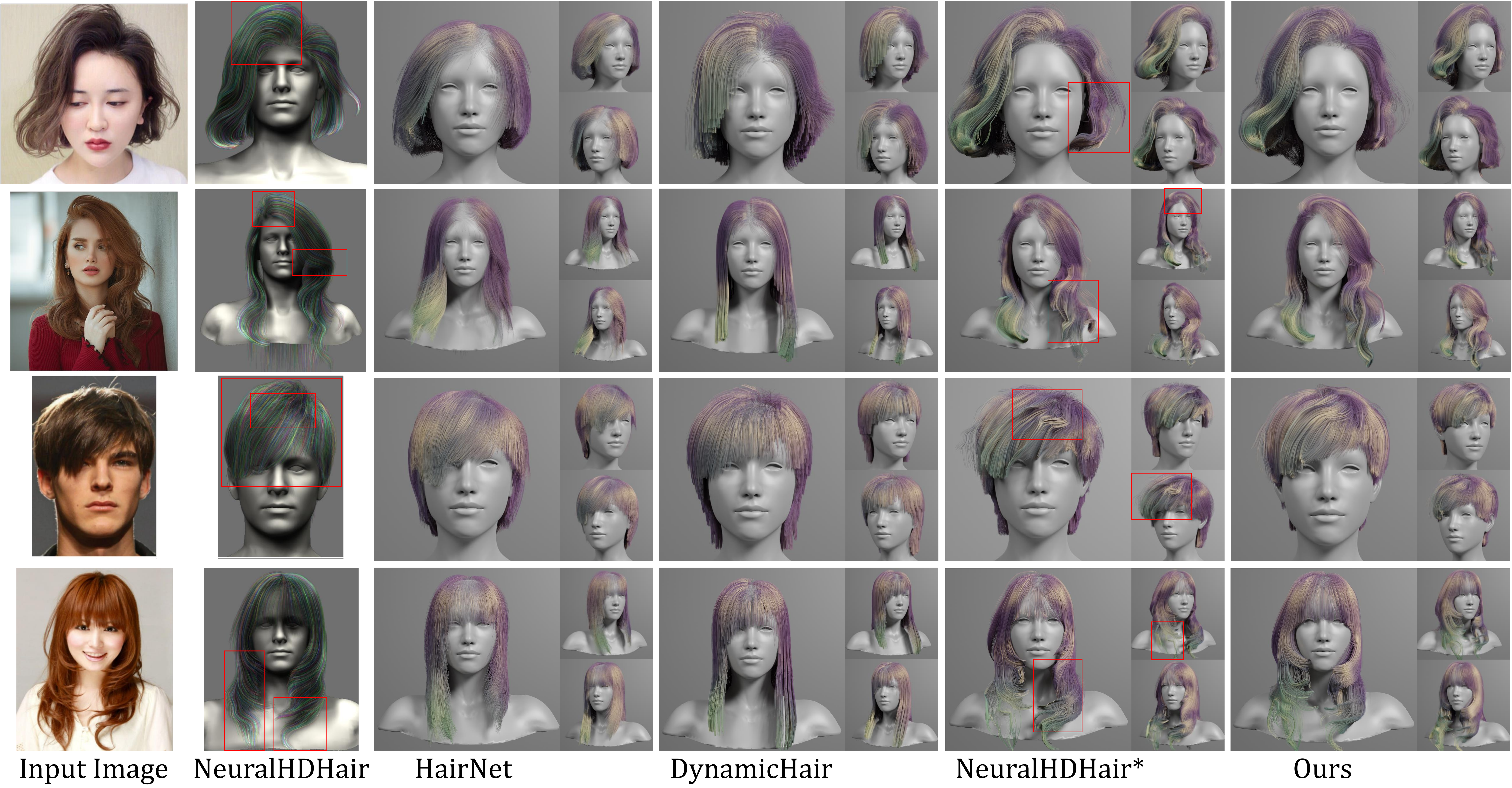}
	\caption{Comparisons with previous methods~\cite{wu2022neuralhdhair,zhou2018hairnet,yang2019dynamic}.}
	\label{fig:compare_neuralhdhair}
\end{figure}

\paragraph{Comparisons on single-view hair modeling.} 
We first compare the reconstruction results of NuralHDHair* (with the input of undirected orientation map), NuralHDHair* with our \repName, HairNet~\cite{zhou2018hairnet}, DynamicHair~\cite{yang2019dynamic} and the original NeuralHDHair~\cite{wu2022neuralhdhair} in~\cref{fig:compare_neuralhdhair}. 
We re-train HairNet and DynamicHair on our synthetic split, as they have not released pre-trained models. 
Based on a global latent code, HairNet and DynamicHair tend to generate coarse shapes while are not capable to reconstruct complex hairstyles. 
With the aid of the voxel-aligned feature and the implicit 3D representation, NeuralHDHair and NeuralHDHair* can produce decent results generally. 
However, it fails in the region with sharp variation of depth and the region with complicated pattern of hair growth (see~\cref{fig:compare_neuralhdhair}). 
The reason could be that the undirected orientation map from Gabor filters can not provide clean and enough information for 3D hair modeling. 
Thanks to the novel representation \repName, our results achieve the best.

\begin{figure}
	\centering
	\includegraphics[width=0.9\linewidth]
	{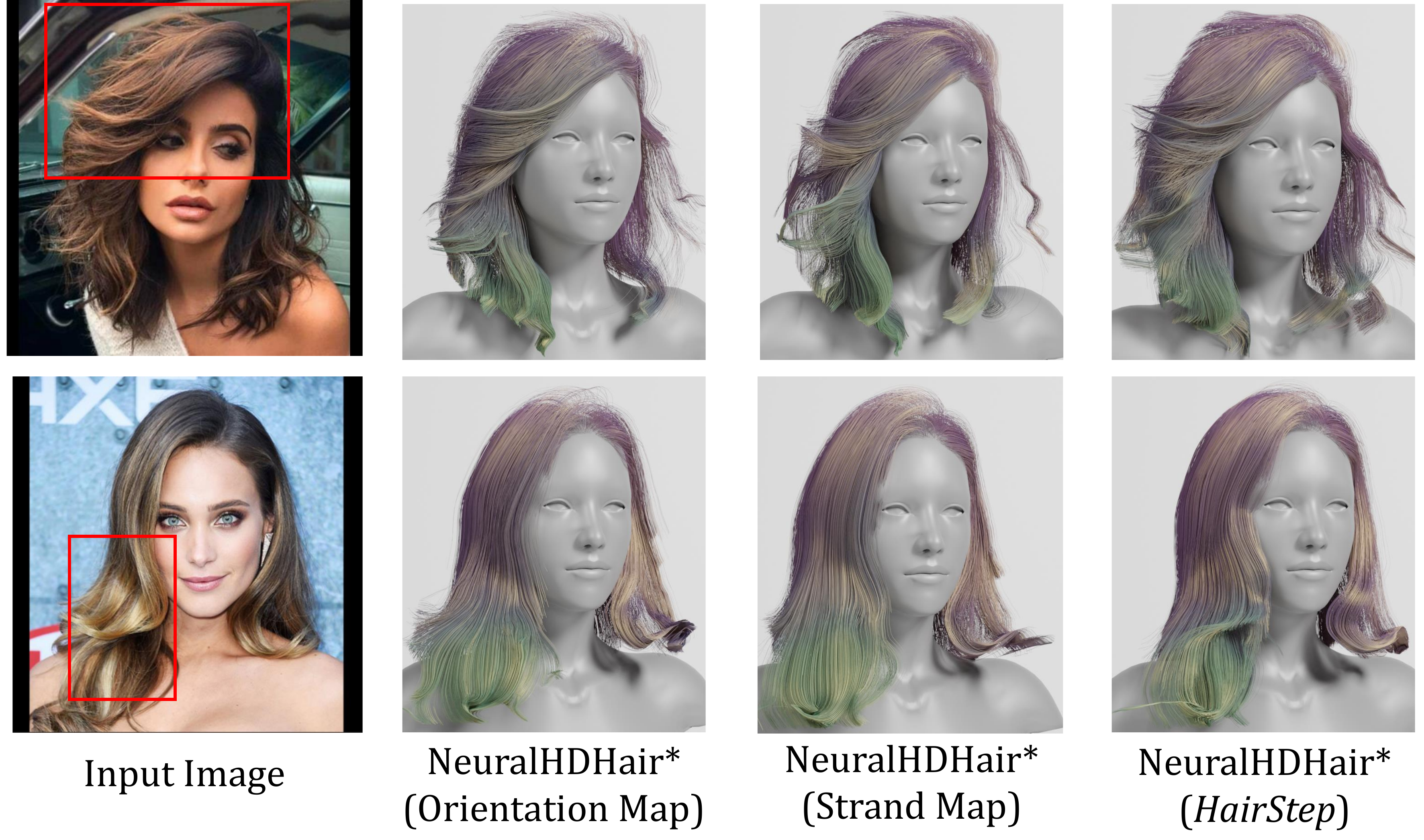}
	\caption{Qualitative evaluation results. From left to right: input images, results of NeuralHDHair*, results using our strand map based representation, and results of our full method, respectively.}
	\label{fig:compare_representation}
\end{figure}

\paragraph{Comparisons on representation.}
To evaluate the effectiveness of our \repName, we compare it with strand map and existing orientation map~\cite{zhou2018hairnet} on three different frameworks, i.e. NeuralHDHair*, DynamicHair~\cite{yang2019dynamic} and HairNet~\cite{zhou2018hairnet}. 
Quantitative comparisons on synthetic and real 
data are illustrated in~\cref{compare:synthetic_compare_existing} and~\cref{compare:real_compare_existing}, respectively. 
As shown in~\cref{compare:synthetic_compare_existing}, our representation benefits all of these three methods on synthetic data. 
Since HairNet can only output explicit hair strands, we follow~\cite{zhou2018hairnet} to report mean square distance error in~\cref{compare:synthetic_compare_existing}. 
In the evaluations on \emph{HiSa} and \emph{HiDa} of~\cref{compare:real_compare_existing}, it is proved that using our strand map achieves better alignment of hair growth than using previous orientation map~\cite{zhou2018hairnet} which suffers from ambiguous direction and image noise. 
The generalization ability of HairNet and DynamicHair is limited by the usage of global feature. 
Hence, directly concatenating depth information to the input does not seem helpful.
Boosting by the full \repName, there is an obvious improvement in depth accuracy on NeuralHDHair*. 
Qualitative comparisons shown in~\cref{fig:compare_representation} yield the same conclusion, where \repName performs the best in depth and preserves fine alignment of hair growth as same as strand map. 
Only applying orientation map leads to undesirable artifacts. 
Note that the depth accuracy of HairNet and DynamicHair in~\cref{compare:real_compare_existing} is based on the low IoU, which is not comparable to NeuralHDHair*. 
In addition, we made a user study on 10 randomly selected examples involving 39 users for reconstructed results of NeuralHDHair* from three representations. 64.87\% chose results from our \repName as the best, while 21.28\% and 13.85\% for strand map and undirected orientation map.

\begin{table}
	\begin{center}
	\small
		\resizebox{\linewidth}{!}{\begin{tabular}{l|c| c}
			\hline
			Method & Orien. err. $\downarrow$ & Occ. acc. $\uparrow$\\
			\hline
		NeuralHDHair* (Orientation map) &  0.1324 & 82.59$\%$ \\
            NeuralHDHair* (Strand map) &  0.0722 \textcolor{blue}{(-41.7\%)} & 84.18$\%$ \\
            NeuralHDHair* (HairStep) & 0.0658 \textcolor{blue}{(-50.3\%)}  & 86.77$\%$ \\
            \hline
            DynamicHair (Orientation map) & 0.1352  & 78.19$\%$   \\
            DynamicHair (Strand map) & 0.1185 \textcolor{blue}{(-12.4\%)}  & 79.62$\%$  \\
            DynamicHair (HairStep) & 0.1174 \textcolor{blue}{(-13.2\%)}  & 79.78$\%$   \\

            \hline
            \hline
            HairNet (Orientation map) & 0.02349  & /  \\
            HairNet (Strand map) & 0.02206 \textcolor{blue}{(-6.1\%)}  & /   \\
            HairNet (HairStep) & 0.02184 \textcolor{blue}{(-7.0\%)}  & /   \\
            \hline
		\end{tabular}}
		\caption{Quantitative comparisons on the USC-HairSalon dataset using different intermediate representations for NeuralHDHair*, DynamicHair~\cite{yang2019dynamic} and HairNet~\cite{zhou2018hairnet}.}
	\label{compare:synthetic_compare_existing}
	\end{center}
\end{table}

\subsection{Ablation Study}
\label{sec:ablation_study}

To better study the effect of each design in depth estimation on the final results, our representation is ablated with three configurations:

\begin{itemize}[leftmargin=*]
    \item \textbf{$\bm{C_{0}}$:} strand map +  $Depth_{pseudo}$.
    \item \textbf{$\bm{C_{1}}$:} strand map + $Depth_{weak}$.
    \item \textbf{$\bm{Full}$:} strand map + $Depth_{DA}$.
\end{itemize}

\begin{figure}
	\centering
	\includegraphics[width=\linewidth]  
	{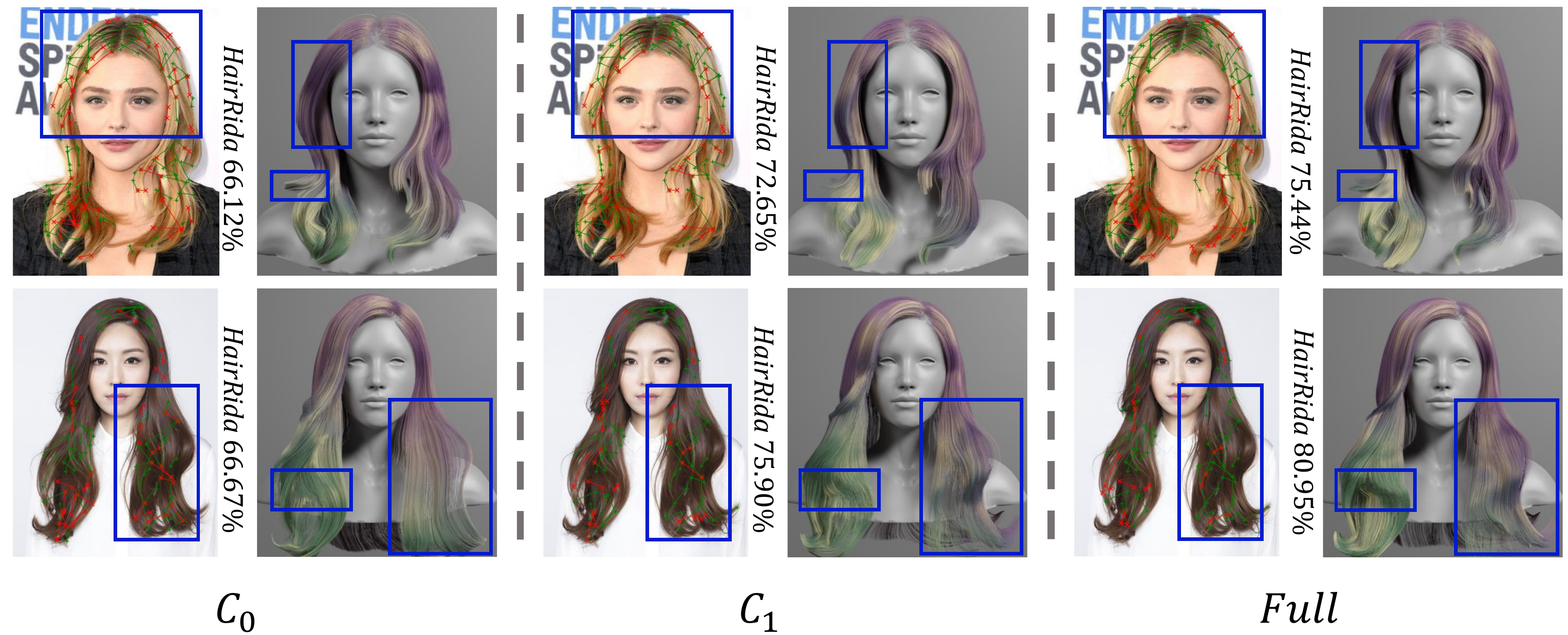}
	\caption{Qualitative ablation study results. Each pair from left to right: input image with the visualization of \depthmetric, where green/red line indicates right/wrong prediction of relative depth of two end point and the reconstructed 3D hair strand model (see \Cref{sec:ablation_study} for detailed explanations).}
	\label{fig:ablation}
\end{figure}

Quantitative comparisons are reported in~\cref{compare:ablation} and qualitative results are shown in \cref{fig:ablation}. 
Our \textbf{$\bm{Full}$} representation achieves the best result in depth accuracy and the decent alignment of hair growth. 
\textbf{$\bm{C_{0}}$} suffers from the flat geometry of depth. 
Meanwhile, \textbf{$\bm{C_{1}}$} can produce results with decent depth accuracy, but obtain a relatively larger difference on the alignment of hair growth than the \textbf{$\bm{Full}$} representation. 



\begin{table}
	\begin{center}
	\small
		\resizebox{1.0\linewidth}{!}{\begin{tabular}{l|c| c| c}
			\hline
			Method & IoU $\uparrow$ & \strandmetric $\downarrow$ & \depthmetric $\uparrow$\\
			\hline
            NeuralHDHair* (Orientation map) &  77.56$\%$ & 19.6  &  70.67$\%$ \\
            NeuralHDHair* (Strand map) &  77.6$\%$ & \textbf{16} \textcolor{blue}{(-18.4\%)}  &  72.37$\%$ \\
            NeuralHDHair* (HairStep) & 77.22$\%$  & 16.36 \textcolor{blue}{(-16.5\%)}  & \textbf{76.79$\%$}  \\
            \hline
            DynamicHair (Orientation map) & 56.39$\%$  & 32.66  & 74.08$\%$  \\
            DynamicHair (Strand map) & 59.51$\%$  & 26.53 \textcolor{blue}{(-18.8\%)}  & 73.42$\%$  \\
            DynamicHair (HairStep) & 59.14$\%$  & 27.51 \textcolor{blue}{(-15.8\%)}  & 73.58$\%$  \\
            \hline
            HairNet (Orientation map) & 57.15$\%$  & 31.97  & 75.65$\%$  \\
            HairNet (Strand map) & 57.48$\%$  & 28.6 \textcolor{blue}{(-10.5\%)} & 74.81$\%$  \\
            HairNet (HairStep) & 57.01$\%$  & 27.68 \textcolor{blue}{(-13.4\%)} & 74.97$\%$  \\
            \hline
\end{tabular}}
		\caption{Quantitative comparisons on \emph{HiSa} and \emph{HiDa} of different intermediate representations for NeuralHDHair*, DynamicHair~\cite{yang2019dynamic} and HairNet~\cite{zhou2018hairnet}.}
	\label{compare:real_compare_existing}
	\end{center}
\end{table}

\begin{table}
	\begin{center}
    \footnotesize
		\begin{tabular}{l|c| c| c}
			\hline
			Method & IoU $\uparrow$ & \strandmetric $\downarrow$ & \depthmetric $\uparrow$\\
            \hline
            \textbf{$\bm{C_{0}}$} & 77.75$\%$  & 16.03 \textcolor{blue}{(-18.2\%)}  & 73.57$\%$  \\
            \textbf{$\bm{C_{1}}$} & 77.11$\%$  & 16.54 \textcolor{blue}{(-15.6\%)}  & 75.8$\%$  \\
            \textbf{$\bm{Full}$} & 77.22$\%$  & 16.36 \textcolor{blue}{(-16.5\%)}  & \textbf{76.79$\%$}  \\
			\hline
		\end{tabular}
		\caption{Quantitative ablation study about depth estimation.}
	\label{compare:ablation}
	\end{center}
\end{table}

\section{Conclusion}
\label{sec:concl}
In this work, we rethink the overall solution of single-view 3D hair modeling and argue that an appropriate intermediate representation for bridging the domain gap between synthetic and real data is essential. 
To this end, we propose a novel 3D hair representation \repName, which consists of a strand map and a depth map, to narrow the existing domain gap. 
We also collect two datasets, i.e., \stranddataset and \depthdataset, with manually annotated strand maps and depth from real portrait images. These datasets not only allow the training of our learning based approach but also introduce fair and objective metrics to evaluate the performance of single-view 3D hair modeling. Extensive experiments on diverse examples demonstrate the effectiveness of our novel representation. 
Our method may fail on some rare and complex hairstyles, because the 3D network is basically overfitted on current synthetic datasets with limited amount and diversity. 


\paragraph{\textbf{Acknowledgement.}}
The work was supported in part by NSFC with Grant No. 62293482, the Basic Research Project No. HZQB-KCZYZ-2021067 of Hetao Shenzhen-HK S$\&$T Cooperation Zone, the National Key R$\&$D Program of China with grant No. 2018YFB1800800, by Shenzhen Outstanding Talents Training Fund 202002, by Guangdong Research Projects No. 2017ZT07X152 and No. 2019CX01X104, by the Guangdong Provincial Key Laboratory of Future Networks of Intelligence (Grant No. 2022B1212010001), and by Shenzhen Key Laboratory of Big Data and Artificial Intelligence (Grant No. ZDSYS201707251409055). It was also partially supported by Shenzhen General Project with No. JCYJ20220530143604010. 




{\small
\bibliographystyle{ieee_fullname}
\bibliography{egbib}

\begin{thebibliography}{10}\itemsep=-1pt

\bibitem{bao2018survey}
Yongtang Bao and Yue Qi.
\newblock A survey of image-based techniques for hair modeling.
\newblock {\em IEEE Access}, 6:18670--18684, 2018.

\bibitem{chai2016autohair}
Menglei Chai, Tianjia Shao, Hongzhi Wu, Yanlin Weng, and Kun Zhou.
\newblock Autohair: Fully automatic hair modeling from a single image.
\newblock {\em ACM Transactions on Graphics}, 35(4), 2016.

\bibitem{chai2013dynamic}
Menglei Chai, Lvdi Wang, Yanlin Weng, Xiaogang Jin, and Kun Zhou.
\newblock Dynamic hair manipulation in images and videos.
\newblock {\em ACM Transactions on Graphics (TOG)}, 32(4):1--8, 2013.

\bibitem{chai2012single}
Menglei Chai, Lvdi Wang, Yanlin Weng, Yizhou Yu, Baining Guo, and Kun Zhou.
\newblock Single-view hair modeling for portrait manipulation.
\newblock {\em ACM Transactions on Graphics (TOG)}, 31(4):1--8, 2012.

\bibitem{chen2016single}
Weifeng Chen, Zhao Fu, Dawei Yang, and Jia Deng.
\newblock Single-image depth perception in the wild.
\newblock {\em Advances in neural information processing systems}, 29, 2016.

\bibitem{eigen2015predicting}
David Eigen and Rob Fergus.
\newblock Predicting depth, surface normals and semantic labels with a common
  multi-scale convolutional architecture.
\newblock In {\em Proceedings of the IEEE international conference on computer
  vision}, pages 2650--2658, 2015.

\bibitem{fan2022self}
Hehe Fan, Xiaojun Chang, Wanyue Zhang, Yi Cheng, Ying Sun, and Mohan
  Kankanhalli.
\newblock Self-supervised global-local structure modeling for point cloud
  domain adaptation with reliable voted pseudo labels.
\newblock In {\em Proceedings of the IEEE/CVF Conference on Computer Vision and
  Pattern Recognition}, pages 6377--6386, 2022.

\bibitem{hoiem2005automatic}
Derek Hoiem, Alexei~A Efros, and Martial Hebert.
\newblock Automatic photo pop-up.
\newblock In {\em ACM SIGGRAPH}, pages 577--584, 2005.

\bibitem{hu2014robust}
Liwen Hu, Chongyang Ma, Linjie Luo, and Hao Li.
\newblock Robust hair capture using simulated examples.
\newblock {\em ACM Transactions on Graphics (TOG)}, 33(4):1--10, 2014.

\bibitem{hu2015single}
Liwen Hu, Chongyang Ma, Linjie Luo, and Hao Li.
\newblock Single-view hair modeling using a hairstyle database.
\newblock {\em ACM Transactions on Graphics (ToG)}, 34(4):1--9, 2015.

\bibitem{isola2017image}
Phillip Isola, Jun-Yan Zhu, Tinghui Zhou, and Alexei~A Efros.
\newblock Image-to-image translation with conditional adversarial networks.
\newblock In {\em Proceedings of the IEEE conference on computer vision and
  pattern recognition}, pages 1125--1134, 2017.

\bibitem{karsch2014depth}
Kevin Karsch, Ce Liu, and Sing~Bing Kang.
\newblock Depth transfer: Depth extraction from video using non-parametric
  sampling.
\newblock {\em IEEE transactions on pattern analysis and machine intelligence},
  36(11):2144--2158, 2014.

\bibitem{ladicky2014pulling}
Lubor Ladicky, Jianbo Shi, and Marc Pollefeys.
\newblock Pulling things out of perspective.
\newblock In {\em Proceedings of the IEEE conference on computer vision and
  pattern recognition}, pages 89--96, 2014.

\bibitem{li2015depth}
Bo Li, Chunhua Shen, Yuchao Dai, Anton Van Den~Hengel, and Mingyi He.
\newblock Depth and surface normal estimation from monocular images using
  regression on deep features and hierarchical crfs.
\newblock In {\em Proceedings of the IEEE conference on computer vision and
  pattern recognition}, pages 1119--1127, 2015.

\bibitem{liang2021domain}
Jian Liang, Dapeng Hu, and Jiashi Feng.
\newblock Domain adaptation with auxiliary target domain-oriented classifier.
\newblock In {\em Proceedings of the IEEE/CVF Conference on Computer Vision and
  Pattern Recognition}, pages 16632--16642, 2021.

\bibitem{lin2018hallucinated}
Kwan-Yee Lin and Guanxiang Wang.
\newblock Hallucinated-iqa: No-reference image quality assessment via
  adversarial learning.
\newblock In {\em Proceedings of the IEEE conference on computer vision and
  pattern recognition}, pages 732--741, 2018.

\bibitem{liu2015deep}
Fayao Liu, Chunhua Shen, and Guosheng Lin.
\newblock Deep convolutional neural fields for depth estimation from a single
  image.
\newblock In {\em Proceedings of the IEEE conference on computer vision and
  pattern recognition}, pages 5162--5170, 2015.

\bibitem{liu2019soft}
Shichen Liu, Tianye Li, Weikai Chen, and Hao Li.
\newblock Soft rasterizer: A differentiable renderer for image-based 3d
  reasoning.
\newblock In {\em Proceedings of the IEEE/CVF International Conference on
  Computer Vision}, pages 7708--7717, 2019.

\bibitem{luo2012multi}
Linjie Luo, Hao Li, Sylvain Paris, Thibaut Weise, Mark Pauly, and Szymon
  Rusinkiewicz.
\newblock Multi-view hair capture using orientation fields.
\newblock In {\em 2012 IEEE Conference on Computer Vision and Pattern
  Recognition}, pages 1490--1497. IEEE, 2012.

\bibitem{luo2013wide}
Linjie Luo, Cha Zhang, Zhengyou Zhang, and Szymon Rusinkiewicz.
\newblock Wide-baseline hair capture using strand-based refinement.
\newblock In {\em Proceedings of the IEEE Conference on Computer Vision and
  Pattern Recognition}, pages 265--272, 2013.

\bibitem{nam2019strand}
Giljoo Nam, Chenglei Wu, Min~H Kim, and Yaser Sheikh.
\newblock Strand-accurate multi-view hair capture.
\newblock In {\em Proceedings of the IEEE/CVF Conference on Computer Vision and
  Pattern Recognition}, pages 155--164, 2019.

\bibitem{newell2016stacked}
Alejandro Newell, Kaiyu Yang, and Jia Deng.
\newblock Stacked hourglass networks for human pose estimation.
\newblock In {\em European conference on computer vision}, pages 483--499.
  Springer, 2016.

\bibitem{paris2004capture}
Sylvain Paris, Hector~M Briceno, and Fran{\c{c}}ois~X Sillion.
\newblock Capture of hair geometry from multiple images.
\newblock {\em ACM transactions on graphics (TOG)}, 23(3):712--719, 2004.

\bibitem{paris2008hair}
Sylvain Paris, Will Chang, Oleg~I Kozhushnyan, Wojciech Jarosz, Wojciech
  Matusik, Matthias Zwicker, and Fr{\'e}do Durand.
\newblock Hair photobooth: geometric and photometric acquisition of real
  hairstyles.
\newblock {\em ACM Trans. Graph.}, 27(3):30, 2008.

\bibitem{ronneberger2015u}
Olaf Ronneberger, Philipp Fischer, and Thomas Brox.
\newblock U-net: Convolutional networks for biomedical image segmentation.
\newblock In {\em International Conference on Medical image computing and
  computer-assisted intervention}, pages 234--241. Springer, 2015.

\bibitem{saito20183d}
Shunsuke Saito, Liwen Hu, Chongyang Ma, Hikaru Ibayashi, Linjie Luo, and Hao
  Li.
\newblock 3d hair synthesis using volumetric variational autoencoders.
\newblock {\em ACM Transactions on Graphics (TOG)}, 37(6):1--12, 2018.

\bibitem{saito2019pifu}
Shunsuke Saito, Zeng Huang, Ryota Natsume, Shigeo Morishima, Angjoo Kanazawa,
  and Hao Li.
\newblock Pifu: Pixel-aligned implicit function for high-resolution clothed
  human digitization.
\newblock In {\em Proceedings of the IEEE/CVF International Conference on
  Computer Vision}, pages 2304--2314, 2019.

\bibitem{saito2020pifuhd}
Shunsuke Saito, Tomas Simon, Jason Saragih, and Hanbyul Joo.
\newblock Pifuhd: Multi-level pixel-aligned implicit function for
  high-resolution 3d human digitization.
\newblock In {\em Proceedings of the IEEE/CVF Conference on Computer Vision and
  Pattern Recognition}, pages 84--93, 2020.

\bibitem{saxena2008make3d}
Ashutosh Saxena, Min Sun, and Andrew~Y Ng.
\newblock Make3d: Learning 3d scene structure from a single still image.
\newblock {\em IEEE transactions on pattern analysis and machine intelligence},
  31(5):824--840, 2008.

\bibitem{shen2020deepsketchhair}
Yuefan Shen, Changgeng Zhang, Hongbo Fu, Kun Zhou, and Youyi Zheng.
\newblock Deepsketchhair: Deep sketch-based 3d hair modeling.
\newblock {\em IEEE transactions on visualization and computer graphics},
  27(7):3250--3263, 2020.

\bibitem{silberman2012indoor}
Nathan Silberman, Derek Hoiem, Pushmeet Kohli, and Rob Fergus.
\newblock Indoor segmentation and support inference from rgbd images.
\newblock In {\em European conference on computer vision}, pages 746--760.
  Springer, 2012.

\bibitem{simonyan2014very}
Karen Simonyan and Andrew Zisserman.
\newblock Very deep convolutional networks for large-scale image recognition.
\newblock {\em arXiv preprint arXiv:1409.1556}, 2014.

\bibitem{song2020learning}
Liangchen Song, Yonghao Xu, Lefei Zhang, Bo Du, Qian Zhang, and Xinggang Wang.
\newblock Learning from synthetic images via active pseudo-labeling.
\newblock {\em IEEE Transactions on Image Processing}, 29:6452--6465, 2020.

\bibitem{su2020robustfusion}
Zhuo Su, Lan Xu, Zerong Zheng, Tao Yu, Yebin Liu, and Lu Fang.
\newblock Robustfusion: Human volumetric capture with data-driven visual cues
  using a rgbd camera.
\newblock In {\em European Conference on Computer Vision}, pages 246--264.
  Springer, 2020.

\bibitem{tan2020michigan}
Zhentao Tan, Menglei Chai, Dongdong Chen, Jing Liao, Qi Chu, Lu Yuan, Sergey
  Tulyakov, and Nenghai Yu.
\newblock Michigan: multi-input-conditioned hair image generation for portrait
  editing.
\newblock {\em ACM Transactions on Graphics (TOG)}, 39(4):95--1, 2020.

\bibitem{tang2019neural}
Sicong Tang, Feitong Tan, Kelvin Cheng, Zhaoyang Li, Siyu Zhu, and Ping Tan.
\newblock A neural network for detailed human depth estimation from a single
  image.
\newblock In {\em Proceedings of the IEEE/CVF International Conference on
  Computer Vision}, pages 7750--7759, 2019.

\bibitem{wu2022neuralhdhair}
Keyu Wu, Yifan Ye, Lingchen Yang, Hongbo Fu, Kun Zhou, and Youyi Zheng.
\newblock Neuralhdhair: Automatic high-fidelity hair modeling from a single
  image using implicit neural representations.
\newblock In {\em Proceedings of the IEEE/CVF Conference on Computer Vision and
  Pattern Recognition}, pages 1526--1535, 2022.

\bibitem{yang2019dynamic}
Lingchen Yang, Zefeng Shi, Youyi Zheng, and Kun Zhou.
\newblock Dynamic hair modeling from monocular videos using deep neural
  networks.
\newblock {\em ACM Transactions on Graphics (TOG)}, 38(6):1--12, 2019.

\bibitem{yang2016deep}
Xiaoshan Yang, Tianzhu Zhang, Changsheng Xu, Shuicheng Yan, M~Shamim Hossain,
  and Ahmed Ghoneim.
\newblock Deep relative attributes.
\newblock {\em IEEE Transactions on Multimedia}, 18(9):1832--1842, 2016.

\bibitem{zhang2018modeling}
Meng Zhang, Pan Wu, Hongzhi Wu, Yanlin Weng, Youyi Zheng, and Kun Zhou.
\newblock Modeling hair from an rgb-d camera.
\newblock {\em ACM Transactions on Graphics (TOG)}, 37(6):1--10, 2018.

\bibitem{zhang2019hair}
Meng Zhang and Youyi Zheng.
\newblock {Hair-GAN: Recovering 3D hair structure from a single image using
  generative adversarial networks}.
\newblock {\em Visual Informatics}, 3(2):102--112, 2019.

\bibitem{zhang2021prototypical}
Pan Zhang, Bo Zhang, Ting Zhang, Dong Chen, Yong Wang, and Fang Wen.
\newblock Prototypical pseudo label denoising and target structure learning for
  domain adaptive semantic segmentation.
\newblock In {\em Proceedings of the IEEE/CVF conference on computer vision and
  pattern recognition}, pages 12414--12424, 2021.

\bibitem{zhang2019ranksrgan}
Wenlong Zhang, Yihao Liu, Chao Dong, and Yu Qiao.
\newblock Ranksrgan: Generative adversarial networks with ranker for image
  super-resolution.
\newblock In {\em Proceedings of the IEEE/CVF International Conference on
  Computer Vision}, pages 3096--3105, 2019.

\bibitem{zhao2019geometry}
Shanshan Zhao, Huan Fu, Mingming Gong, and Dacheng Tao.
\newblock Geometry-aware symmetric domain adaptation for monocular depth
  estimation.
\newblock In {\em Proceedings of the IEEE/CVF Conference on Computer Vision and
  Pattern Recognition}, pages 9788--9798, 2019.

\bibitem{zheng2018t2net}
Chuanxia Zheng, Tat-Jen Cham, and Jianfei Cai.
\newblock T2net: Synthetic-to-realistic translation for solving single-image
  depth estimation tasks.
\newblock In {\em Proceedings of the European conference on computer vision
  (ECCV)}, pages 767--783, 2018.

\bibitem{zhou2018hairnet}
Yi Zhou, Liwen Hu, Jun Xing, Weikai Chen, Han-Wei Kung, Xin Tong, and Hao Li.
\newblock Hairnet: Single-view hair reconstruction using convolutional neural
  networks.
\newblock In {\em Proceedings of the European Conference on Computer Vision
  (ECCV)}, pages 235--251, 2018.

\end{thebibliography}
}

\clearpage

\appendix 
\renewcommand{\appendixname}{Supplementary Material~\Alph{section}}

\renewcommand\thesection{\Alph{section}}
\renewcommand{\thetable}{S\arabic{table}}  
\renewcommand{\thefigure}{S\arabic{figure}}

\centerline{\textbf{\LARGE{-- {Supplementary Material} --}}}

\section{Dataset}
\paragraph{Statistics of data distribution.}
To construct \stranddataset and \depthdataset, we collect 1,250 clear portrait images with various hairstyles from the Internet, where 80\% are female and 20\% are male. 
We classify the collected hairstyles into three classes, i.e., Short, Middle and Long, according to the position of their hair ends. 
If hair ends are above the mouth, the hairstyle will be classified as Short. 
If hair ends are below the shoulder, the hairstyle belongs to Long class.
Otherwise, it is Middle. 
We collect 300 Short hair, 300 Middle hair and 650 Long hair.
As for the curl type, the number of straight, wavy and curly are 210, 620, 420, respectively.

\begin{figure*}[htpb]
	\centering
	\includegraphics[width=0.8\textwidth]  
	{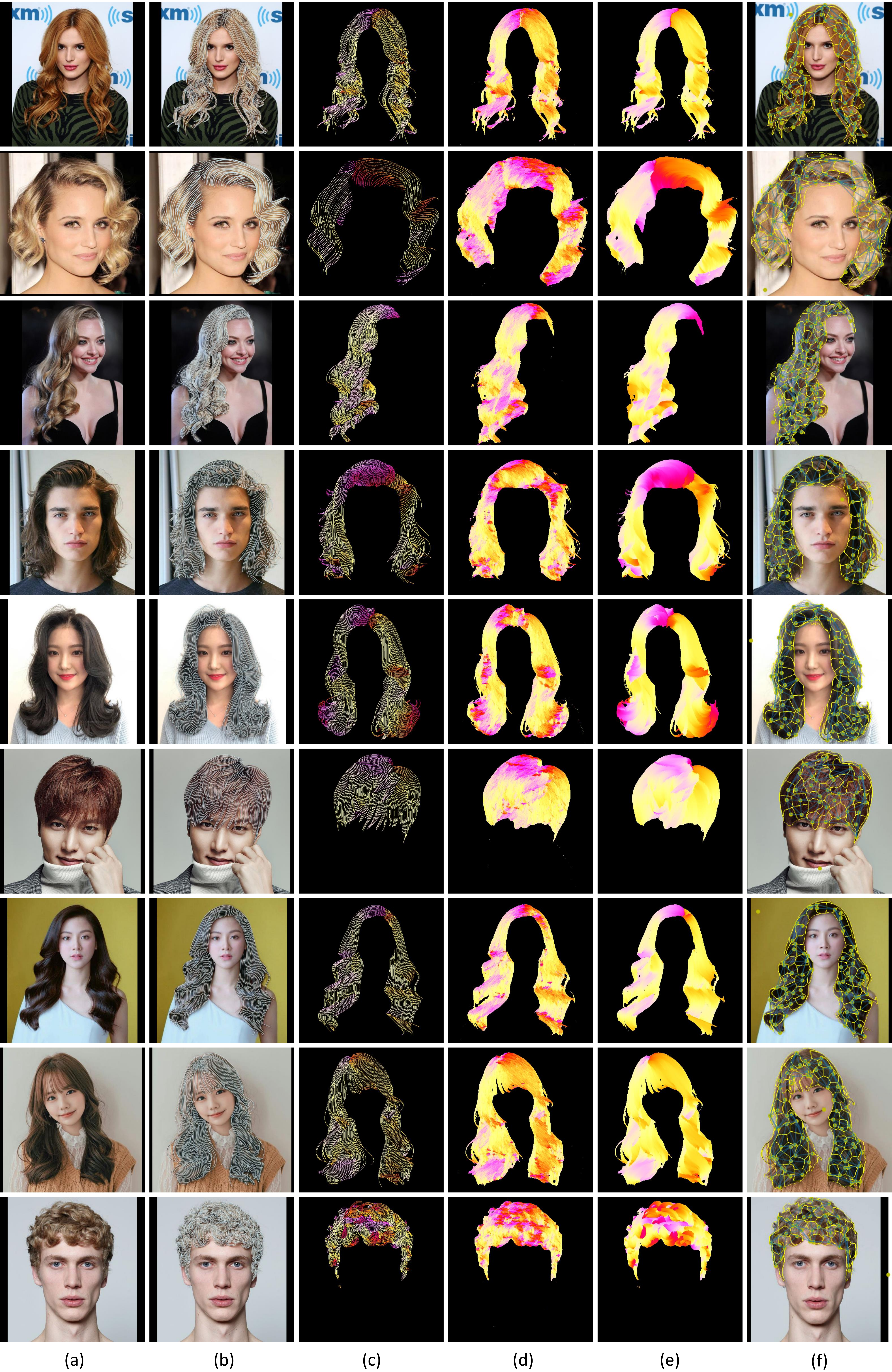}
	\caption{More examples for strand map annotation and depth pair sampling. From left to right: (a) collected images, (b) strokes drawn by artists, (c) colored strokes, (d) undirected orientation maps from Gabor filters, (e) strand maps and (f) super-pixels for depth pair sampling.}
	\label{fig:dataset}
\end{figure*}

More examples about strand map annotation and depth pair sampling are shown in~\cref{fig:dataset}.

\section{Implementation Details}
We describe the details of our networks and training for \repName extraction and 3D hair reconstruction in this section. 

\paragraph{\repName extraction.}
We use the same U-Net in~\cite{isola2017image} to extract strand maps from real images with the resolution of $512\times 512$. 
The network consists of an eight-layer encoder and an eight-layer decoder, where each layer downsamples/upsamples by a factor of 2 and skip connections are adopted between symmetric layers. 
We refer the readers to~\cite{isola2017image} for detailed designs. 
Training is conducted using a batch size of 16 for 50 epochs on 1 NVIDIA RTX3090Ti card for about 12 hours. 
The learning rate is 0.0003. 
During training, the loss weight $\alpha$ is set to 0.1.

We use the same Hourglass network in~\cite{chen2016single} to estimate depth maps for real images with the resolution of $512\times 512$. 
The hourglass network is formed with four stacks, which consists of a series of convolutions, downsampling, upsampling and skip connections. 
Please refer to~\cite{chen2016single} for details.
The network is trained with a batch size of 8 for 100 epochs on 2 NVIDIA RTX3090Ti cards for about 6 hours. 
The learning rate is 0.0003 and the loss weight $\beta$ is set to 0.1. 

\paragraph{3D Hair Reconstruction.}
We use the same structure as the IRHairNet in~\cite{wu2022neuralhdhair}, where we first extract a $96\times 128 \times 128 \times 64$ feature volume from the input representation resized to $256\times 256$ via a U-Net combined with VIFu, then query coarse 3D occupancy field and orientation field with two MLPs. 
As for the fine module, we substitute the luminance map to the input representation resized to $1024\times 1024$ and extract high-resolution occupancy field and orientation field via an hourglass network and two MLPs. 
Please refer to~\cite{wu2022neuralhdhair} for the details of network design. 
We follow~\cite{zhou2018hairnet} to combine the body mask to the mask channel of the strand map/orientation map rather than introducing a new channel.
Note that our \repName has one more depth channel than orientation map and strand map. Thus, the first layers of the encoders have 4 channels when using \repName, while 3 channels when taking the strand map or orientation map as the input. 
Training is conducted using a batch size of 2 for 100 epochs on one NVIDIA RTX3090Ti cards for roughly 5-6 day. The learning rate is initialized set to be 0.0001, and decayed by a factor of 0.1 in the $60_{th}$ epoch. 

\section{Back views}
\begin{figure*}[t]
	\centering
	\includegraphics[width=\linewidth]  
	{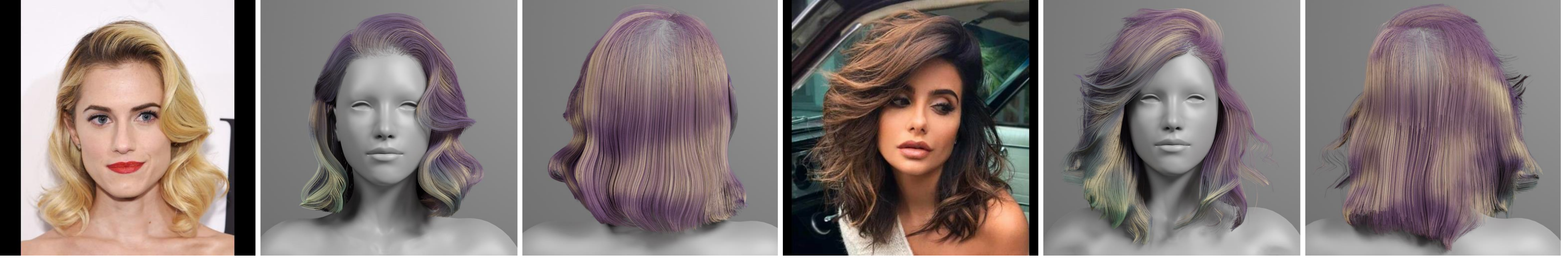}
	\caption{Examples with front and back view.}
     \label{fig:back_view}
\end{figure*}

Two examples of the back view are shown in~\cref{fig:back_view} where the invisible parts tend to be smooth but still reasonable. 
This is because the 3D hair dataset provides shape priors. 

\section{Failure cases}
\begin{figure*}[htpb]
	\centering
	\includegraphics[width=0.6\linewidth]  
	{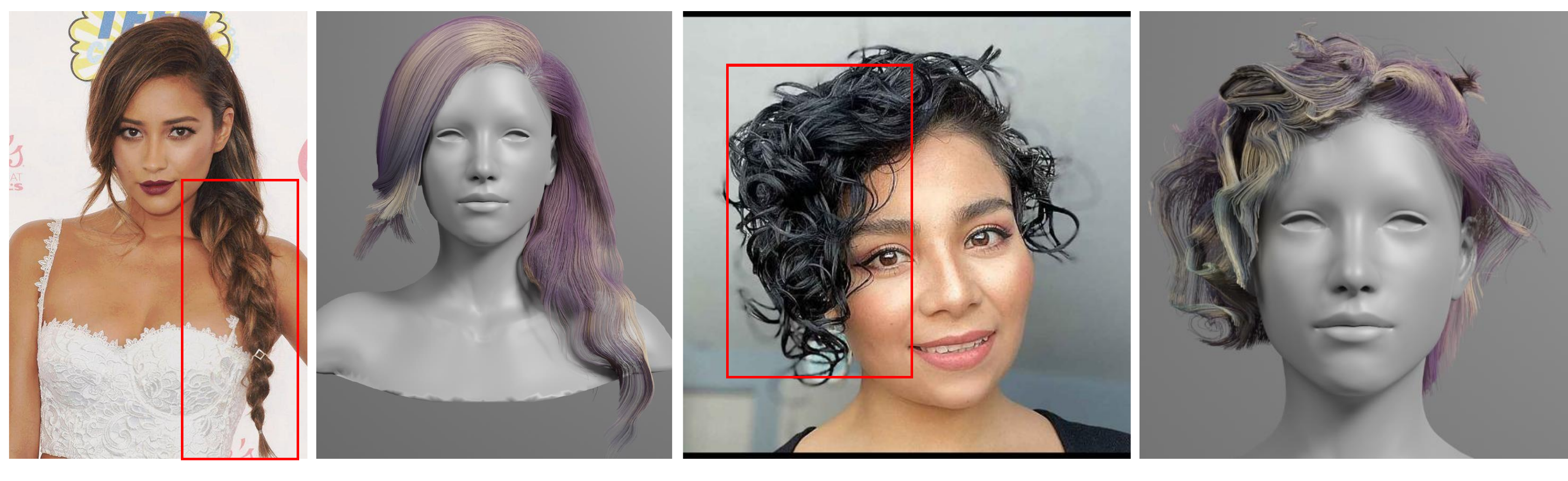}
	\caption{Failure cases.}
     \label{fig:failure_case}
\end{figure*}

As mentioned in the Conclusion, our method may fail on some rare and complex hairstyles, because the existing 3D hair datasets are with limited amount and diversity. 
For example, as shown in~\cref{fig:failure_case}, our method does not work on hairstyles with braid (left) and complex curly pattern (right). 

\section{More Comparisons}

\begin{figure*}[t]
	\centering
	\includegraphics[width=\linewidth]  
	{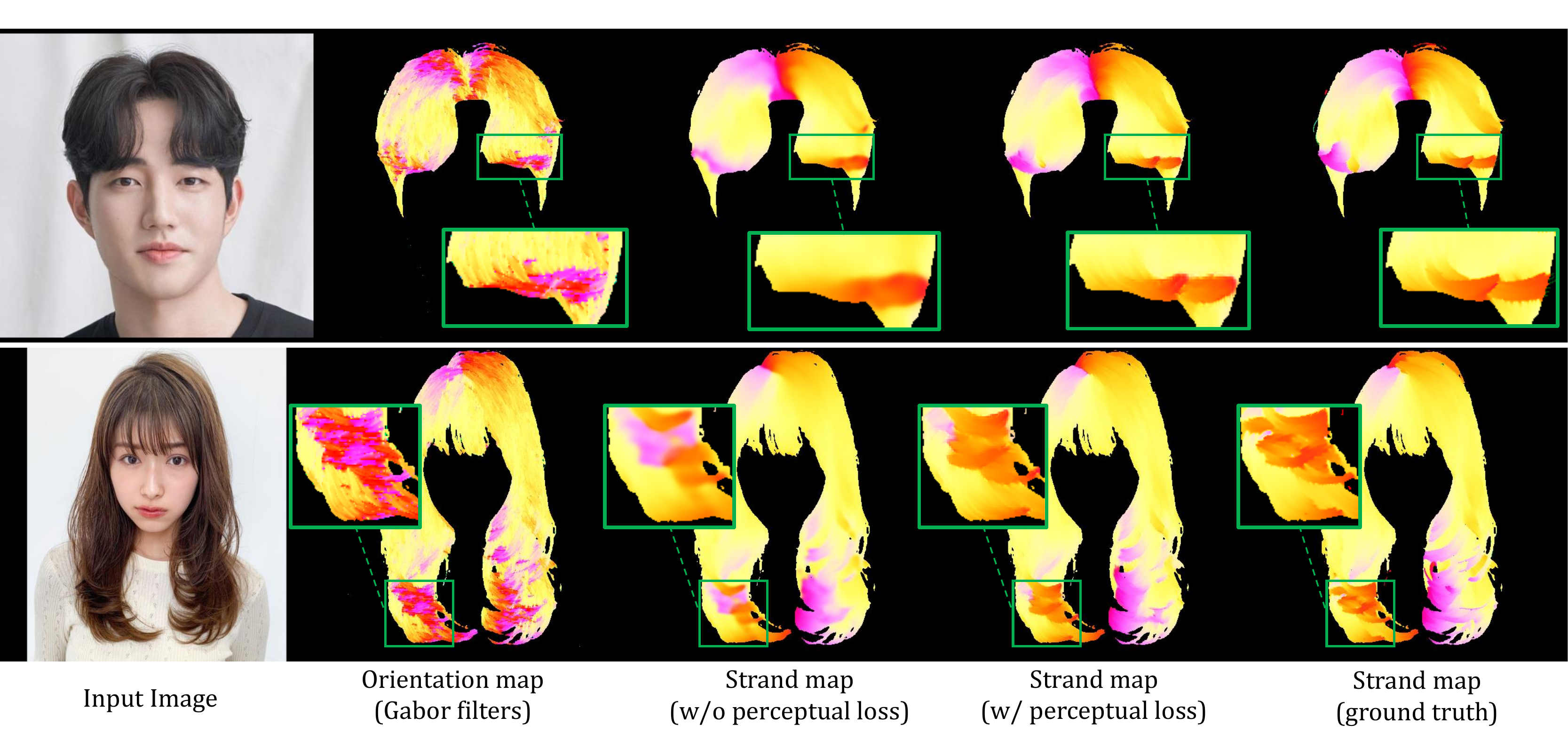}
	\caption{Qualitative comparisons on orientation/strand maps.}
	\label{fig:compare_orien_p}
\end{figure*}

\paragraph{Perceptual loss.} 
We think the perceptual loss is necessary in strand map prediction. 
Although it cannot provide obvious quantitative improvement (w/ 14.2 \textit{v.s.} w/o 14.1), it brings visually sharper local features (\cref{fig:compare_orien_p}).  
Also, we made an extra experiment of 3D hair reconstruction on our method without perceptual loss. We found its \strandmetric and \depthmetric (16.51 and 75.3\%) are worse than using perceptual loss (16.36 and 76.79\%).

\begin{figure*}
	\centering
	\includegraphics[width=\linewidth]
	{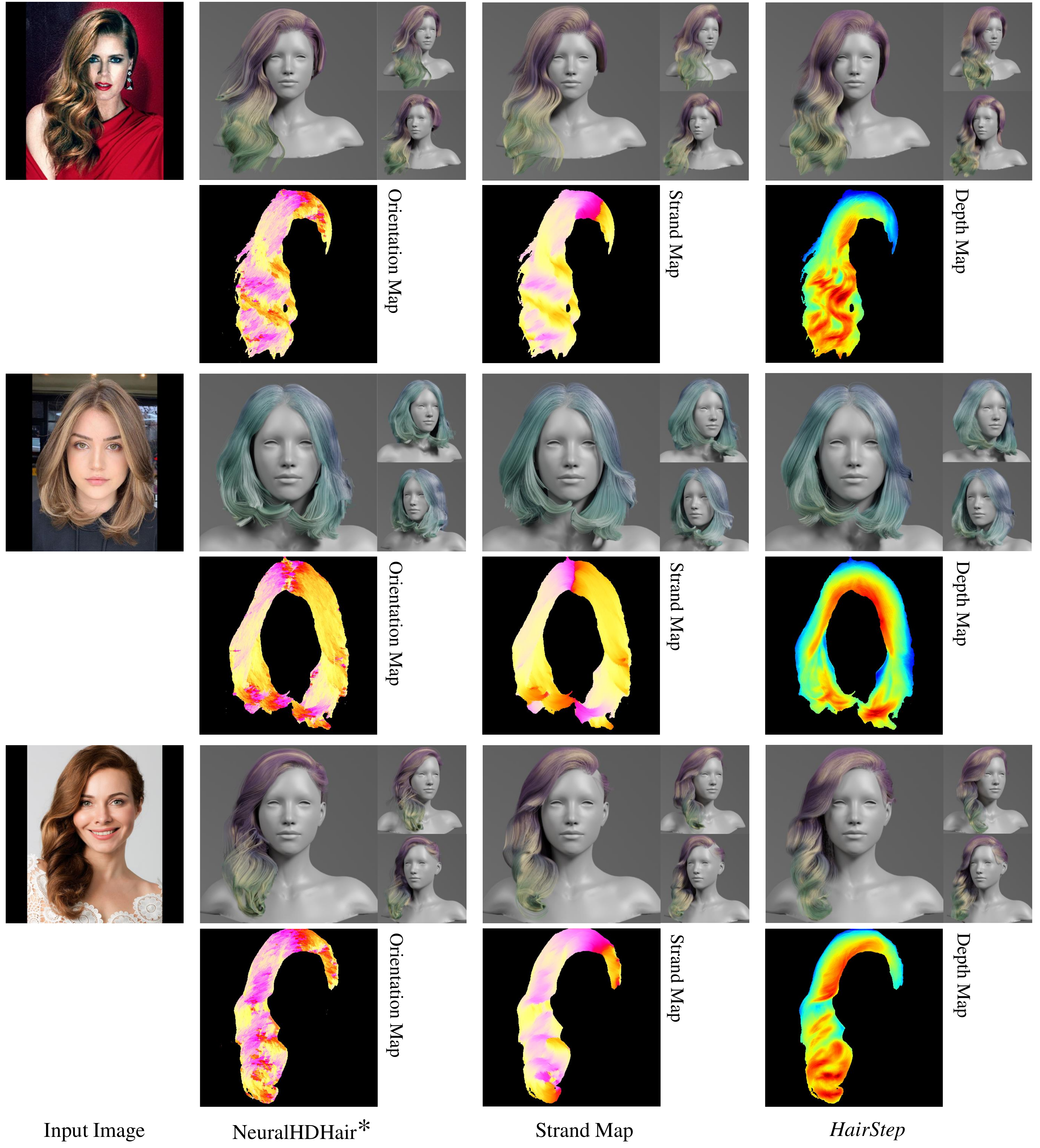}
 
	\caption{More qualitative comparisons for different representations. From left to right: input images, results of NeuralHDHair*, results using our strand map based representation, and results of our full method, respectively. Orientation maps from Gabor filters, predicted strand maps and depth maps are also shown under the reconstructed results.}
	\label{fig:compare_representation_1}
\end{figure*}

\paragraph{Qualitative comparisons for different representations.} 
More qualitative comparisons for different representations are shown in~\cref{fig:compare_representation_1} where using \repName achieves the best results. 

\begin{figure*}
	\centering
	\includegraphics[width=0.95\linewidth]
	{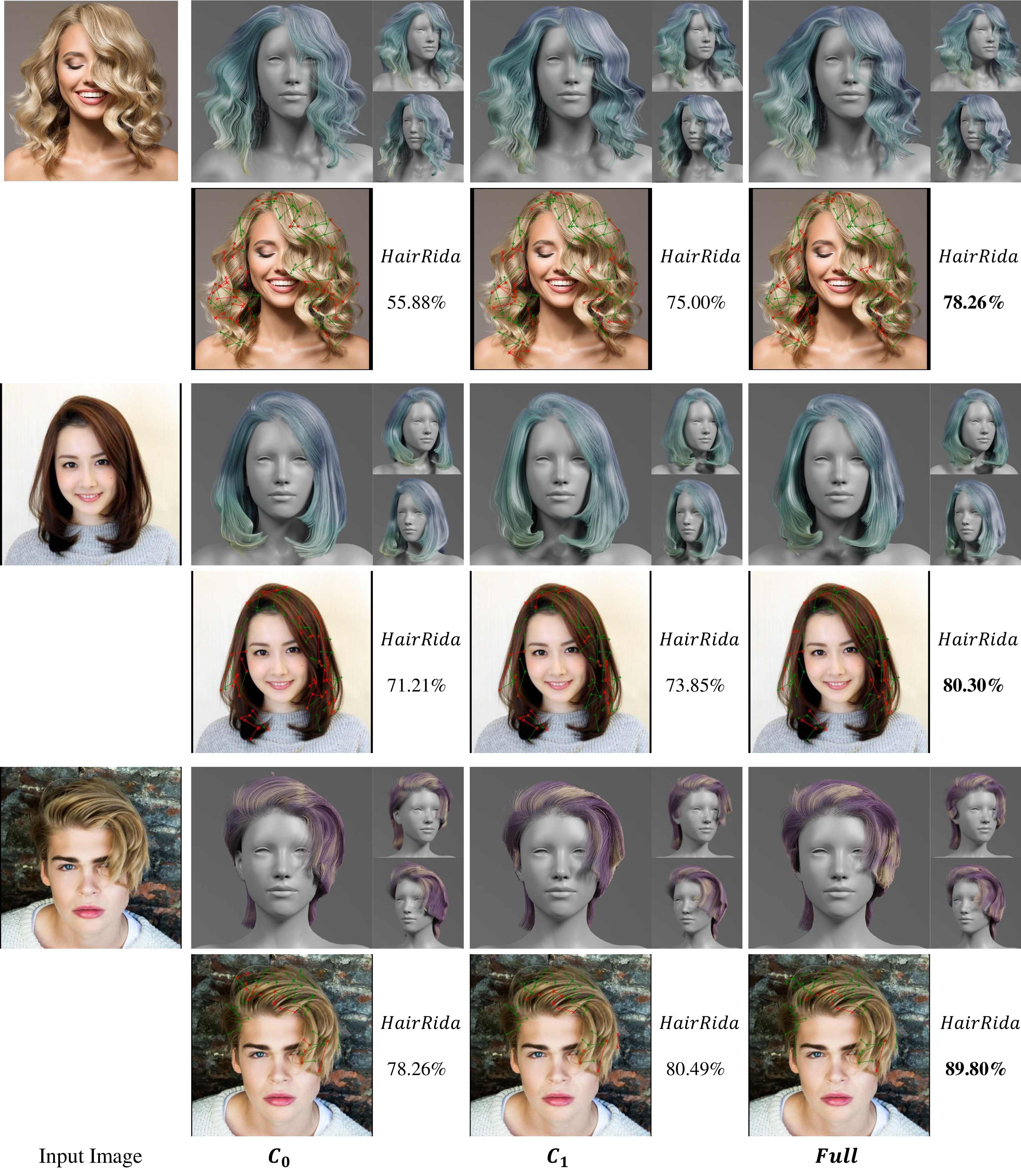}
 \caption{More qualitative comparisons for depth ablation. From left to right: input images, results of $\bm{C_{0}}$, $\bm{C_{1}}$ and our full method. We also visualize the \depthmetric below each reconstructed result, where green/red lines indicate right/wrong predictions.}
	\label{fig:compare_ablation_1}
\end{figure*}

\paragraph{Qualitative comparisons for depth ablation.}
More qualitative comparisons for depth ablation are shown in~\cref{fig:compare_ablation_1} where our full model achieves the best accuracy in depth.


\section{User Study}
\begin{figure*}
	\centering
	\includegraphics[width=\linewidth]
	{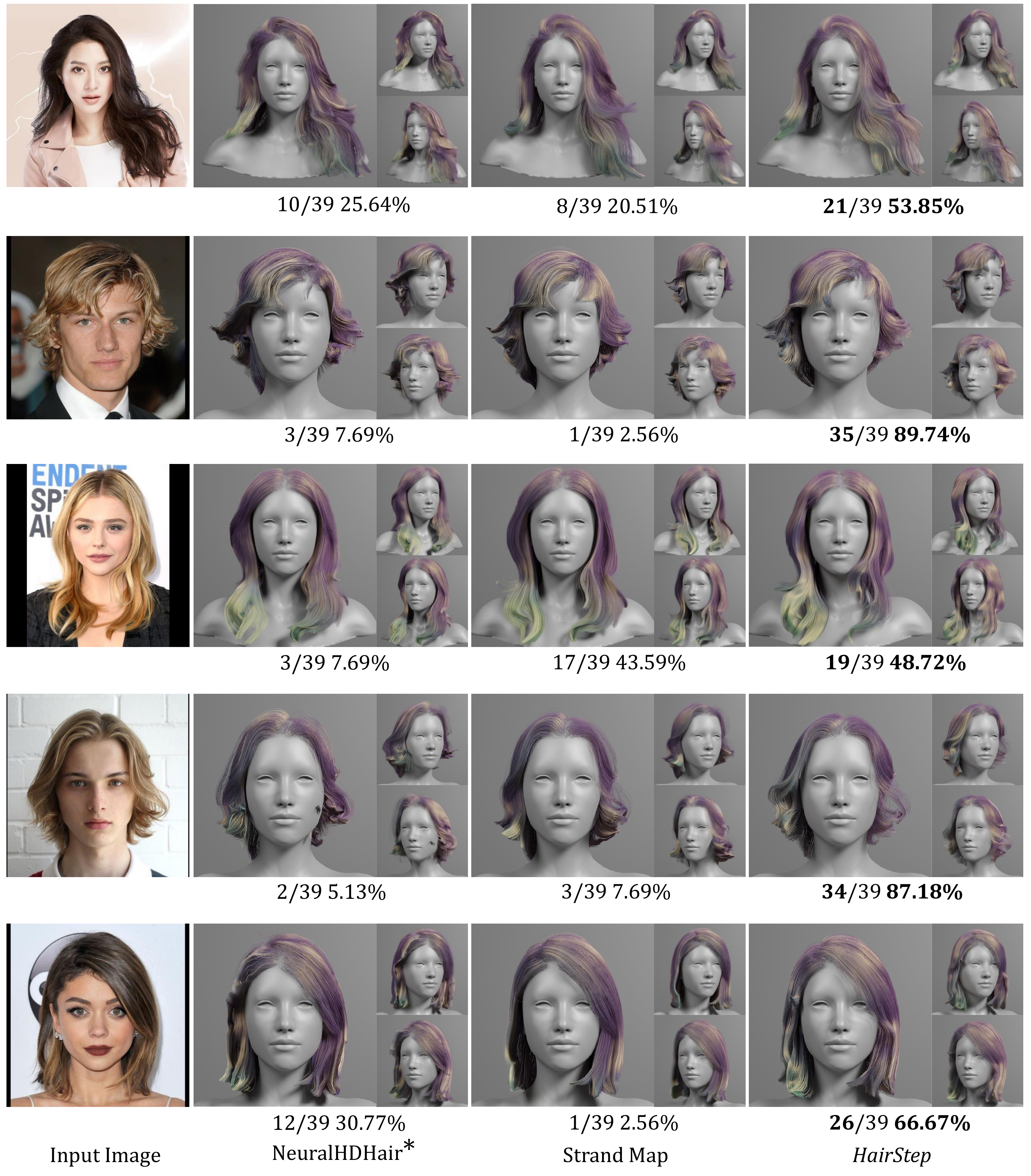}
	\caption{Examples for user study. From left to right: input images, results of NeuralHDHair*, results using our strand map based representation, and results of our full method, respectively. We also provide the statistics of 3 different representations for each example.}
	\label{fig:supply_userstudy_1}
\end{figure*}

\begin{figure*}
	\centering
	\includegraphics[width=\linewidth]
	{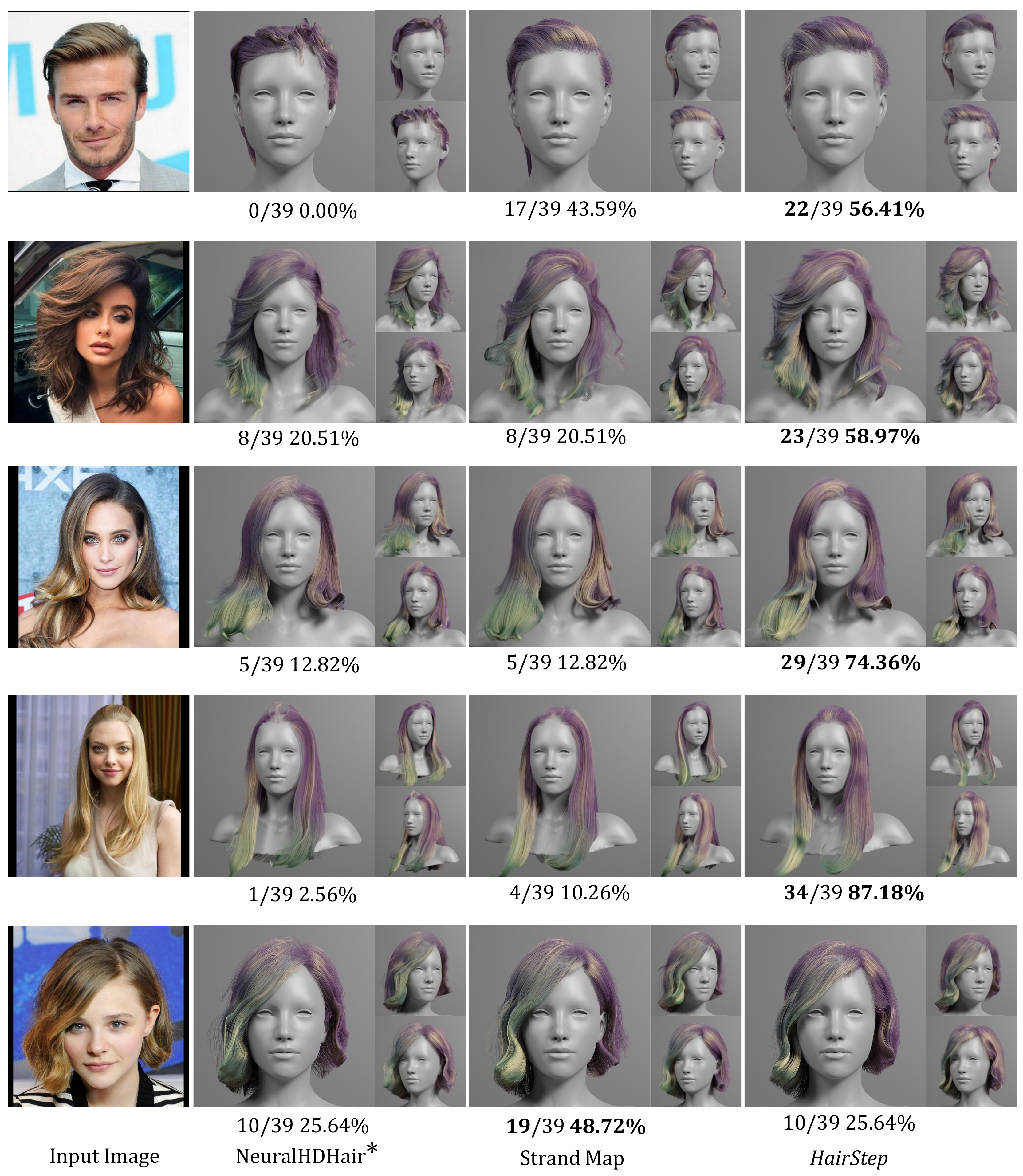}
	\caption{Examples for user study. From left to right: input images, results of NeuralHDHair*, results using our strand map based representation, and results of our full method, respectively. We also provide the statistics of 3 different representations for each example.}
	\label{fig:supply_userstudy_2}
\end{figure*}

We made a user study on 10 randomly selected examples involving 39 users for reconstructed results of NeuralHDHair* from three representations. 64.87\% chose results from our \repName as the best, while 21.28\% and 13.85\% for strand map and undirected orientation map.~\cref{fig:supply_userstudy_1} and~\cref{fig:supply_userstudy_2} provide the statistics of 3 different representations for each example.

\end{document}